\documentclass{article}

\usepackage{PRIMEarxiv}
\usepackage[utf8]{inputenc}
\usepackage[T1]{fontenc}
\usepackage[round,authoryear]{natbib}
\usepackage{hyperref}
\usepackage{url}
\usepackage{booktabs}
\usepackage{amsmath,amssymb,amsfonts}
\usepackage{nicefrac}
\usepackage{microtype}
\usepackage{graphicx}
\usepackage{fancyhdr}
\usepackage{longtable}
\usepackage{threeparttable}
\usepackage[noend]{algpseudocode}
\usepackage{algorithm}
\usepackage{algorithmicx}
\usepackage{textcomp}
\usepackage{makecell}
\usepackage{bbding}
\usepackage{pifont}
\usepackage{subcaption}
\usepackage{tikz}
\usepackage{pgfplots}
\usepackage{xcolor}
\usepackage{float}
\usepackage{multirow}
\pgfplotsset{compat=1.18}
\graphicspath{{Graph/}}

\title{IntElicit: Eliciting and Assessing Contextualized Creativity via Dialogue Policy Optimization}

\author{
\begin{tabular}{c}
Mingjia Li$^{1,\dagger}$, Jin Wu$^{1,\dagger}$, Hong Qian$^{1,2,*}$, Wenhao Huang$^{1}$, Yiyang Huang$^{1}$ \\
Yiwen Zhang$^{1}$, Chanjin Zheng$^{1}$, Xiangfeng Wang$^{1}$, Aimin Zhou$^{1,2}$, Jiajun Guo$^{1}$ \\[0.5em]
$^{1}$East China Normal University \\
$^{2}$Shanghai Innovation Institute\\[0.3em]
$^{\dagger}$Equal contribution. $^{*}$Corresponding author: \texttt{hqian@cs.ecnu.edu.cn}
\end{tabular}
}

\begin{document}
\maketitle

\begin{abstract}Contextualized assessment offers high ecological validity for evaluating creativity but introduces a critical challenge: observed performance may be confounded with cognitive proficiency (domain knowledge) and agency (willingness to engage). Meanwhile, in the age of generative AI, creative problem solving increasingly occurs in tool-mediated and human--AI interactive environments, making fully static assessment less aligned with contemporary creative practice. To address these issues, this paper proposes \textit{IntElicit}, a framework for eliciting and assessing contextualized creativity via dialogue policy optimization. IntElicit functions as a constrained adaptive \textit{AI Interviewer}: it provides non-directive knowledge and agency scaffolds in multi-turn interaction to reduce non-creative confounders, while preserving participants' responsibility for generating the creative content being evaluated. Specifically, to tackle sparse rewards and potential reward hacking (e.g., answer dictation) in open-ended educational dialogue, IntElicit introduces a decomposed process reward mechanism. This mechanism aligns the policy with pedagogical elicitation, rewarding prompts that draw out participant reasoning rather than producing optimal answers on their behalf. Extensive experiments, including participant simulation and a human subject study ($N=64$), show that IntElicit improves elicited creative outcomes over expert-designed baselines. Together, the results suggest that interactive elicitation can reveal creative potential that static FPSP-style assessment may miss, providing a formative and diagnostic lens for contextualized creativity assessment in AI-mediated learning contexts.
\end{abstract}

\section{Introduction}
\label{sec:introduction}

Creativity, often defined as the ability to generate products that are both novel and useful, is widely recognized as a critical competency for the 21st century \citep{gluaveanu2010paradigms, taguma2019oecd, llego202221st}. As artificial intelligence increasingly automates routine cognitive tasks, the human capacity for creative problem-solving has become central to education and workforce development \citep{brynjolfsson2014second}. However, despite its importance, assessing creativity in educationally meaningful contexts remains difficult because creative performance is shaped not only by idea generation, but also by background knowledge, confidence, engagement, and the interactional conditions under which students are asked to respond. This challenge echoes broader concerns that traditional assessments often provide only discrete snapshots of performance and may be insufficiently adapted to learners' backgrounds and contemporary AI-mediated practices~\citep{swiecki2022assessment}.

Traditional approaches to creativity assessment largely fall into two categories: self-report scales and performance tests. Self-report measures often suffer from social subjectivity bias, where individuals may misjudge their abilities due to factors like the Dunning-Kruger effect~\citep{Kruger1999Unskilled} or social desirability~\citep{paulhus1984two, silvia2012assessing}. Performance tests, conversely, attempt to measure creativity objectively but vary significantly in their contextual fidelity. Common approaches range from simple tasks like the Alternative Uses Task (AUT)~\citep{runco2012divergent} and Realistic Presented Problems (RPP)~\citep{chand1993problem}, to highly contextualized frameworks such as the Future Problem Solving Program (FPSP)~\citep{FPSP1, FPSP2, FPSP3}. While AUT and RPP are widely used, their simplistic instructions (e.g.,``list uses for a brick'') suffer from \textit{limited ecological validity}~\citep{baer2015domain, zeng2011can}, failing to capture the complex nature of real-world problem-solving.

To address this ecological gap, our work aligns with the paradigm of FPSP, which evaluates participants within immersive, realistic scenarios. However, while FPSP improves contextual fidelity, a fully static implementation still faces two limitations. First, performance in complex scenarios can be entangled with cognitive and agential factors, such as domain knowledge, task comprehension, confidence, and willingness to elaborate~\citep{runco1995cognition}. A participant may therefore perform poorly not because they lack creative potential, but because they fail to externalize that potential under unsupported testing conditions. Second, in the age of generative AI, creative problem solving increasingly occurs in tool-mediated and human--AI collaborative environments, where people clarify problems, explore alternatives, and refine ideas through interaction~\citep{rezwana2023designing,noy2023experimental,oliveira2025assessing}. From an authentic assessment perspective, the assessment situation should resemble the kinds of practices in which the target competence is expected to be used. Interactive elicitation addresses these two limitations by making participants' reasoning more observable under adaptive, non-directive scaffolding and by modeling an assessment format closer to contemporary AI-mediated creative work. Consequently, we argue that \textit{interactive elicitation} is a necessary extension of contextualized creativity assessment.

\begin{figure}
\centering
  \includegraphics[width=0.9\textwidth]{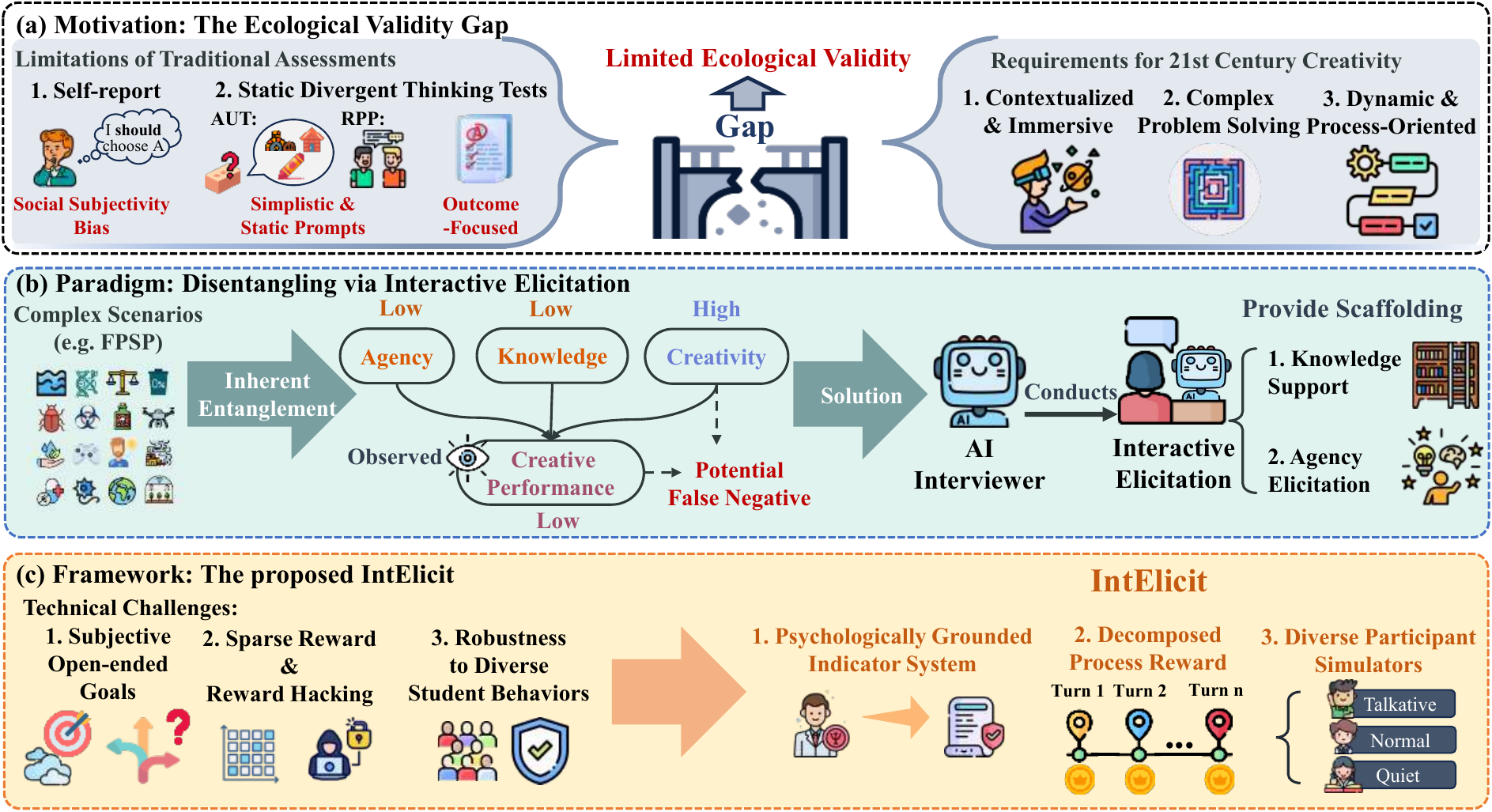}
  \caption{Research motivation and overview of the proposed IntElicit framework.
  \textbf{(a)} The \textit{Ecological Validity Gap}: Static creativity assessments may miss dynamic, process-oriented reasoning in contextualized problem-solving tasks.
\textbf{(b)} The \textit{Interactive Elicitation Paradigm}: An AI interviewer provides adaptive scaffolding (e.g., knowledge support and agency elicitation) to reduce non-creative confounders during assessment while preserving the participant's responsibility for generating ideas.
\textbf{(c)} The \textit{IntElicit Architecture}: The framework combines multidimensional creativity indicators, decomposed process rewards, and diverse participant simulators to support open-ended dialogue policy optimization.}
  \label{fig:intro}
\end{figure}

As illustrated in Figure~\ref{fig:intro}, our approach first motivates interactive elicitation as a response to the ecological validity gap, and then implements it as an adaptive AI interviewer framework for contextualized creativity assessment. To realize this vision, we propose \textit{IntElicit}, an \textbf{Int}eractive \textbf{Elicit}ation framework powered by Dialogue Policy Optimization that functions as \textit{an adaptive AI Interviewer}. Importantly, the role of the AI is not to co-author the creative response or train participants to become more creative during the task. Instead, IntElicit acts as a constrained assessment scaffold: it may help participants clarify task contexts, sustain engagement, elaborate their reasoning, and reflect on alternatives, but the creative content to be assessed must remain participant-generated. While Large Language Models (LLMs) offer a promising foundation for such an AI Interviewer~\citep{achiam2023gpt,kasneci2023chatgpt}, optimizing them for this role presents significant technical challenges. The primary hurdle is optimizing dialogue policies for non-verifiable, open-ended goals. Unlike domains with verifiable rewards such as math or coding~\citep{ouyang2022training,lightman2023let,rafailov2023direct}, creativity assessment is subjective and process-sensitive. Furthermore, in a multi-turn assessment, the reward signal is often sparse (received only at the end). This sparsity can lead to ``reward hacking'', where the AI Interviewer, aiming to maximize the final creativity score, might simply dictate high-quality ideas to the participant rather than eliciting them, which defeats the purpose of assessment. Finally, the agent must be robust, capable of adapting to diverse participant behaviors, from the ``reticent'' interviewee needing encouragement to the ``divergent'' one wandering off-topic.

IntElicit addresses these challenges through a synergistic approach. First, 16 immersive assessment scenarios are designed by expert psychologists following the FPSP paradigm. Grounded in this context, a multidimensional indicator system is constructed to operationalize contextualized creative performance. To prevent reward hacking and support non-directive elicitation, we introduce a decomposed process reward mechanism based on expert pedagogical strategies (e.g., rewarding prompts that encourage participants to identify problems, justify ideas, and reflect on alternatives). We train the policy using a diverse participant simulator, creating a controlled environment populated with simulated participants exhibiting different engagement patterns.

The contributions of this paper are as follows. First, we introduce \textit{interactive elicitation} as a formative and diagnostic paradigm for contextualized creativity assessment, aiming to reduce false negatives caused by knowledge gaps or low agency while preserving the participant's role as the source of creative ideas. Second, we propose \textit{IntElicit}, a dialogue policy optimization framework that learns adaptive assessment scaffolds for open-ended, multi-turn creativity tasks. Third, we introduce a \textit{Decomposed Process Reward} mechanism that rewards pedagogically meaningful elicitation and discourages answer dictation. Finally, through simulated participants, qualitative edge-case analysis, and a human subject study with 64 participants, we show that IntElicit elicits higher-quality creative outputs and adapts to diverse participant behaviors such as reticence and digression.

\section{Related Work}
\label{sec:related_work}

\subsection{Creativity Assessment}
Creativity assessment has traditionally relied on psychometric instruments focusing on Divergent Thinking (DT). The most widely adopted paradigms include the Alternative Uses Task (AUT)~\citep{runco2012divergent}, Realistic Presented Problems (RPP)~\citep{chand1993problem} and the Torrance Tests of Creative Thinking (TTCT)~\citep{torrance1966torrance}. These tests typically employ static prompts (e.g., ``list unusual uses for a brick'') and evaluate responses based on fluency, flexibility, and originality. With the advent of computational linguistic technology including LLMs, recent work has sought to automate the scoring of these tests using semantic distance metrics~\citep{beaty2021automating} or by prompting LLMs as evaluators~\citep{luchini2025automated, organisciak2023beyond, kern2024assessing}, significantly improving assessment efficiency. For a comprehensive review of automated creativity assessment, please refer to~\citep{bahgautomated}.

However, critics argue that traditional DT tests lack \textbf{ecological validity}, as they divorce creative thinking from the complex, domain-specific contexts found in real-world problem solving~\citep{zeng2011can, baer2015domain}. To bridge this gap, Realistic Presented Problems (RPP)~\citep{chand1993problem} and Situational Judgment Tests (SJTs)~\citep{herde2019situational} were introduced to simulate more practical scenarios. The Future Problem Solving Program (FPSP)~\citep{FPSP1, FPSP2, FPSP3} represents a significant advancement in this direction, employing immersive, multi-stage futuristic scenarios to evaluate participants' ability to identify challenges and propose innovative solutions within a constrained narrative.

Despite the high fidelity of FPSP-style assessments, they face a critical challenge when automated: the \textbf{confounding of creativity with cognitive and agential factors}~\citep{runco1995cognition}. A participant's failure to produce a creative solution may stem from a lack of domain knowledge, task comprehension, confidence, or willingness to elaborate rather than a lack of creative potential. Human interviewers can partially address this issue by dynamically clarifying the task, encouraging elaboration, and redirecting attention to relevant scenario constraints. However, existing computational methods typically treat creativity assessment as a static input-output mapping, leaving these non-creative confounders unaddressed. This limitation becomes more salient as creative problem solving increasingly occurs in AI-mediated environments, where interaction with intelligent tools is part of authentic practice~\citep{rezwana2023designing,noy2023experimental}. Our work therefore proposes an interactive elicitation framework that acts as a constrained AI interviewer: it reduces the influence of non-creative confounders through multi-turn knowledge and agency scaffolding, while preserving the participant as the source of the creative ideas being evaluated.

\subsection{Multi-turn Dialogue Optimization}
Optimizing dialogue agents for purposeful interaction has evolved from supervised fine-tuning (SFT) to Reinforcement Learning from Human/AI Feedback (RLHF/RLAIF)~\citep{ouyang2022training,  DBLP:journals/corr/abs-2204-05862, DBLP:journals/corr/abs-2212-08073, DBLP:conf/icml/0001PMMFLBHCRP24}. In complex tasks, standard RLHF often struggles with sparse rewards, where feedback is only available at the end of a long conversation~\citep{lightman2023let, DBLP:conf/nips/WuHSDSASOH23, DBLP:conf/naacl/AmmanabroluULSR21}. To address this, recent research focuses on process supervision and reward decomposition. \citet{DBLP:journals/corr/abs-2505-15922} proposed aligning agents with global feedback via reward decomposition, effectively breaking down long-term goals into dense, turn-level signals. Similarly, \citet{DBLP:journals/corr/abs-2508-03905} introduced Sotopia-RL, which optimizes social intelligence in multi-turn interactions by designing rewards that capture the nuances of social goals and information exchange.

Furthermore, the role of dialogue agents is shifting from passive information retrieval to active collaboration. CollabLLM~\citep{DBLP:conf/icml/WuGP0LDC0L025} highlights the importance of transitioning agents from passive responders to active collaborators capable of initiating structure and guiding joint problem-solving. In the educational domain, this aligns with the concept of instructional scaffolding, where agents provide temporary support to guide learners~\citep{DBLP:conf/icalt/DaiLJLTGC23}. However, applying these optimization techniques to creativity assessment presents a unique \textbf{\textit{alignment tax}}~\citep{ouyang2022training}: the agent must elicit the participant's ideas without dictating the answer (reward hacking). Unlike math or coding tasks with verifiable solutions~\citep{lightman2023let, DBLP:journals/corr/abs-2503-06639}, creative elicitation requires a delicate balance between guidance and open-endedness. IntElicit bridges this gap by employing a decomposed process reward specifically designed to penalize idea-dictation while rewarding the elicitation of participant agency and knowledge application.

\section{Problem Formulation of Interactive Creativity Elicitation}

Our primary objective is to develop an adaptive AI interviewer capable of eliciting the creative potential of participants with diverse personality traits. We formulate this task as a dialogue policy optimization problem within an open-ended environment. Unlike traditional tasks with immediate feedback, creativity elicitation is characterized by sparse and subjective rewards, where the quality of ideas is only observable after the interaction concludes.

Formally, we model the interaction environment using a set of diverse LLM-based simulated participants, denoted as $U$. The interaction unfolds over $n$ turns, generating a trajectory $\mathcal{T} = \{\tau_1, \dots, \tau_n\}$. Each turn consists of a pair $\tau_i = (\tau_{i,A}, \tau_{i,U})$, where $\tau_{i,A}$ is the utterance generated by the AI interviewer's policy $\pi_\theta(\tau_{i,A} | \mathcal{T}_{<i})$ given the history $\mathcal{T}_{<i} = \{\tau_1, \dots, \tau_{i-1}\}$, and $\tau_{i,U}$ is the subsequent response from the simulated participant $u \sim U$.

Upon the completion of the assessment, the environment yields a multi-dimensional sparse reward vector $[R_{\text{Nove}}, R_{\text{Comp}}, R_{\text{Flex}}, R_{\text{Appr}}]$, corresponding to four creativity dimensions, namely, \textit{Novelty}, \textit{Complexity}, \textit{Flexibility}, and \textit{Appropriateness}, respectively. The reward vector is determined by $\mathbf{R}(\cdot)$: an LLM-as-a-Judge mechanism grounded in a psychological indicator system.

To optimize the policy $\pi_\theta$, we define a scalar evaluation function $\text{AHP}(\cdot)$ that aggregates these dimensions via a weighted sum, where the specific coefficients are derived using the Analytic Hierarchy Process (AHP)~\citep{saaty1980analytic} detailed in Section~\ref{sec:decomposition}. The optimization objective is to find the parameters $\theta^*$ that maximize the expected creative performance over the distribution of participants:
\begin{equation}
    \theta^* = \mathop{\arg\max}_{\theta} \mathbb{E}_{u \sim U, \tau \sim \pi_\theta} \left[ \text{AHP}(\mathbf{R}(\mathcal{T})) \right]\,.
\end{equation}

\section{The Proposed IntElicit Framework}
\begin{figure}
\centering
  \includegraphics[width=0.9\textwidth]{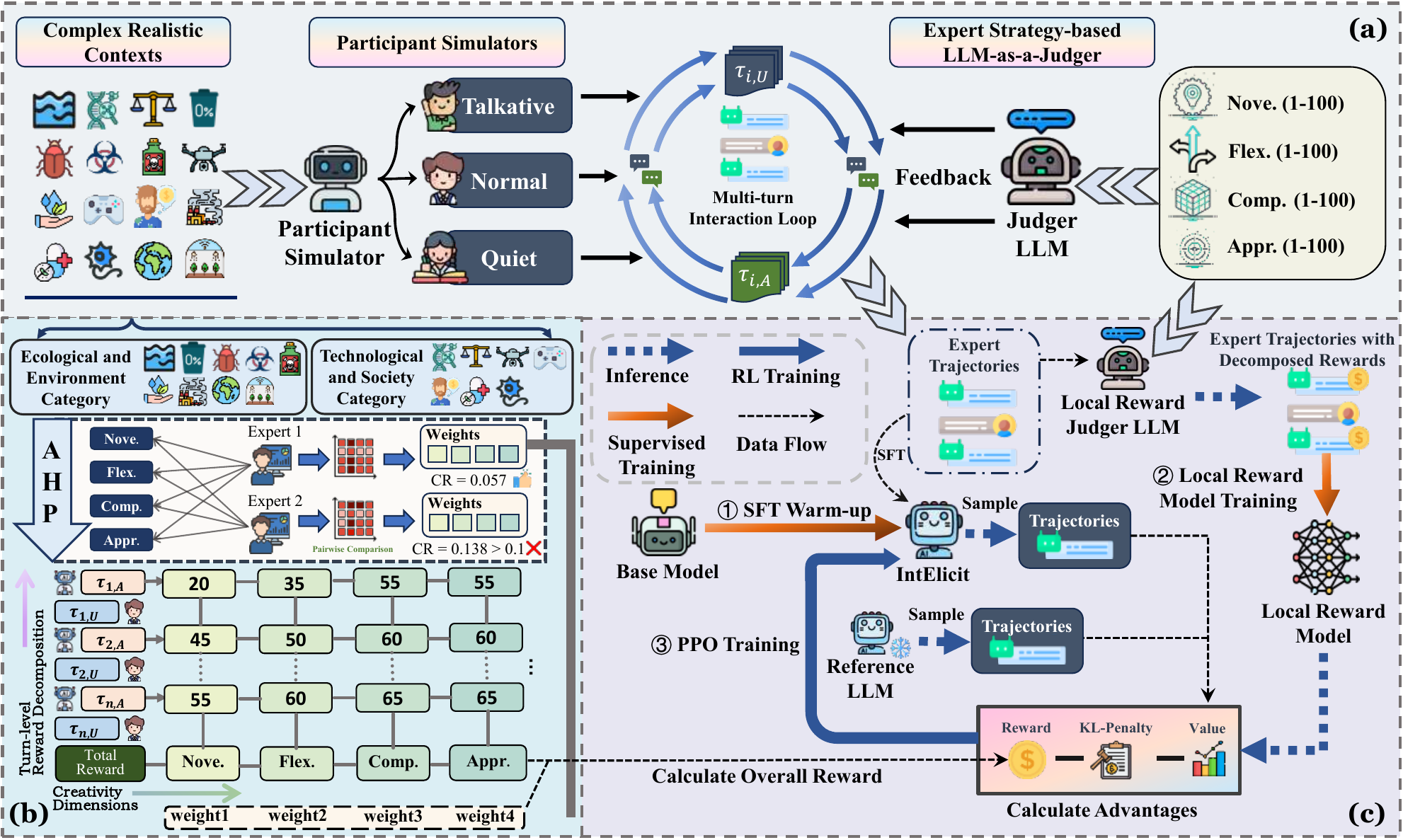}
  \caption{Schematic architecture of the IntElicit training pipeline.
\textbf{(a) Data generation}: Participant simulators with different engagement personas interact with the interviewer across expert-designed contextualized scenarios to generate multi-turn trajectories.
\textbf{(b) Reward construction}: Forward interaction sampling and an expert-guided LLM-as-a-Judge produce decomposed process rewards, with scenario-level dimension weights derived through AHP.
\textbf{(c) Policy optimization}: The interviewer policy is first initialized with supervised fine-tuning on expert trajectories and then optimized with PPO guided by the learned local reward model.}
  \label{fig:framework}
\end{figure}

This section presents IntElicit, a framework designed to adaptively foster participant creativity through dialogue. Our approach begins with an \textit{Interactive Forward Sampling}, where an LLM-based interviewer performs multi-turn rollouts with diverse simulated participants across complex realistic scenarios. An LLM-as-a-Judge evaluator grounded in expert psychological guidelines is then employed to evaluate these interactions. Subsequently, we introduce the \textit{Turn-level Reward Decomposition} mechanism to deconstruct holistic feedback into fine-grained signals corresponding to specific interaction dimensions. Finally, we discuss the \textit{Dialogue Policy Optimization} strategy, which leverages these decomposed rewards to enable the interviewer to adaptively elicit the creative potential of participants with diverse personas. An overview of IntElicit is shown in Figure~\ref{fig:framework}.
\subsection{Participant Simulator}
To simulate realistic interactions, IntElicit defines a simulator function $\mathcal{S}_u$ for each specific participant persona $u \in U$. Formally, this simulator is modeled as a mapping function $\mathcal{S}_u: (\mathcal{T}_{<i}, \tau_{i,A}) \to \tau_{i,U}$, which yields the participant response $\tau_{i,U}$ based on the history $\mathcal{T}_{<i}$ and the current interviewer utterance $\tau_{i,A}$. We operationalize this simulation by injecting expert-curated prompts into an LLM, conditioning it to generate utterances that strictly adhere to the designated personas, linguistic styles, and behavioral archetypes associated with the persona $u$. This ensures that the simulator consistently maintains its character constraints throughout the dialogue, providing a robust environment for training the adaptive interviewer.

The simulator is used as a training and stress-testing environment, not as a substitute for human validation. Its main function is to expose the interviewer policy to repeated interactional variation that would be prohibitively expensive to collect from human participants at scale. Each persona prompt specifies the participant's response length, willingness to elaborate, level of initiative, and tendency to remain on topic. During interaction, the simulator receives the full dialogue history and the current interviewer utterance, which allows it to produce context-sensitive replies rather than independent single-turn responses. This design makes the policy optimization problem closer to an educational interview: the interviewer must adapt to the participant's previous reasoning, decide when to ask for elaboration, and avoid collapsing into answer dictation.

\subsection{Forward Interaction Sampling}
\label{sec:fowardsampling}
In this framework, we employ a Monte Carlo-based forward sampling approach, extending conversations turn-by-turn until a termination criterion is met (either reaching the maximum turn limit or the completion of the interaction). To construct optimal interaction trajectories, we must evaluate and select the best response generated by the interviewer at each turn $i$. However, computing a full-trajectory reward for every candidate is computationally prohibitive. Therefore, IntElicit introduces a hyperparameter of look-ahead window size $w$ (look ahead $w$ turns)  to constrain the forward simulation. This strategy significantly reduces computational overhead while retaining sufficient contextual information for accurate evaluation. The process is detailed as follows:

\paragraph{Candidate Response Generation.} At the $i$-th interaction turn, given the dialogue history $\mathcal{T}_{<i} = \{\tau_1, \dots, \tau_{i-1}\}$ and the current participant utterance $\tau_{i,U}$, we perform parallel sampling of $K$ candidate responses from the interviewer policy $\pi_\theta$: $\{\tau_{i,A}^{(k)}\}_{k=1}^K \sim \pi_{\theta}( \cdot | \mathcal{T}_{<i}, \tau_{i,U})$.

\paragraph{Forward Interaction Sampling.} For each candidate response $\tau_{i,A}^{(k)}$, we sample $N$ future trajectories via alternating interactions between the participant simulator $\mathcal{S}_u$ and the interviewer policy $\pi_\theta$. Specifically, for a participant with persona $u \in U$, the simulation follows the iterative process:
\begin{align}
\tau_{i+l,U}^{(k)} &= \mathcal{S}_u(\mathcal{T}_{<i+l}^{(k)}, \tau_{i+l,A}^{(k)})\,, \\
\tau_{i+l+1,A}^{(k)} &\sim \pi_\theta(\cdot \mid \mathcal{T}_{<i+l}^{(k)} \cup \{\tau_{i+l,U}^{(k)}\})\,,
\end{align}
where $l= 0,1,\ldots,w-1$. Each sampled trajectory segment is defined as $\mathcal{T}_{i:i+w}^{(k)} = \{\tau_{i}^{(k)}, \dots, \tau_{i+w}^{(k)}\}$.

\paragraph{Optimal Response Selection.} We compute the multi-dimensional reward vector for each sampled trajectory, the overall reward is then aggregated using AHP. Finally, we select the candidate with the highest expected score as the output for the current turn $\tau_{i,A}^*=\tau_{i,A}^{(\hat{k})}$ where $\hat{k}=\arg\max_{k}\text{AHP}(\mathbf{R}(\mathcal{T}_{i:i+w}^{(k)}))$.

\subsection{Turn-level Reward Decomposition}
\label{sec:decomposition}

In open-ended creative dialogues, relying solely on episode-level sparse rewards presents two significant challenges. First, the \textbf{sparsity} of the signal makes policy optimization notoriously difficult, as the agent receives no feedback on intermediate steps essential for guiding the conversation. Second, optimizing for a final outcome often leads to \textbf{reward hacking}. For instance, an interviewer might dictate a high-quality solution to quickly terminate the task. While this maximizes the outcome score, it stifles the participant's proactive inquiry and prevents the emergence of original, self-generated insights. To mitigate these issues, we propose a mechanism that decomposes the holistic objective into fine-grained, turn-level supervision.

\paragraph{Scenario-Adaptive Weighting via AHP.}
\label{sec:ahp}
To prevent the policy from collapsing into narrow optimization, we first identify four critical dimensions of creativity, represented as the reward vector $[R_{\text{Nove}}, R_{\text{Flex}}, R_{\text{Comp}}, R_{\text{Appr}}]$. However, the importance of these dimensions varies significantly across contexts. To derive rational weights, we employ the Analytic Hierarchy Process (AHP), engaging two expert psychologists to prioritize these dimensions within two distinct scenario categories: \textit{Ecological-Environment} and \textit{Technological-Society}. The consistency of these judgments was verified using the Consistency Ratio (CR), ensuring $\text{CR} < 0.1$. Based on the derived priority vectors, we obtain scenario-specific weights $\boldsymbol{\lambda}^{(c)}$, normalized such that $\sum_{j} \lambda_j^{(c)} = 1$. Consequently, the final scalar training objective $R_{\text{total}}$ for a trajectory $\mathcal{T}$ is formulated as the weighted dot product: 
\begin{equation} 
R_{\text{total}}(\mathcal{T}) = \text{AHP}(\mathbf{R}(\mathcal{T})) = \boldsymbol{\lambda}^{(c)} \cdot \mathbf{R}(\mathcal{T}) \,.
\end{equation}
This mechanism allows the reward function to dynamically balance divergent facets based on scenario requirements. For instance, as detailed in our expert calibration (Table~\ref{tab:ahp_weights}), the system prioritizes \textbf{Complexity} in \textit{Ecological} contexts to address the systemic interconnectedness of environmental issues, while emphasizing \textbf{Novelty} in \textit{Technological} scenarios to encourage innovative solution generation.

\begin{table}[ht]
\centering
\small
\caption{AHP-derived weights for the four creativity dimensions in the two scenario categories.}
\label{tab:ahp_weights}
\begin{tabular}{l|cccc}
\toprule
\textbf{Scenario Category} & \textbf{Novelty} & \textbf{Complexity} & \textbf{Appropriateness} & \textbf{Flexibility} \\
\midrule
Ecological and Environment & 0.1416 & 0.3089 & 0.2407 & 0.3089 \\
Technological and Society & 0.3125 & 0.0625 & 0.3125 & 0.3125 \\
\bottomrule
\end{tabular}
\end{table}

\paragraph{Turn-level Process Decomposition.}
While AHP addresses ``what'' to evaluate regarding the final outcome, it does not address ``when'' to reward effective scaffolding. A truly creative spark often requires a strategic build-up rather than emerging immediately after a prompt. Relying solely on sparse, outcome-based rewards often fails to capture these intermediate pedagogical achievements. To address this, we implement a \textbf{Turn-level Process Decomposition} mechanism. Specifically, we employ an LLM-as-a-Judge equipped with specific \textit{process-oriented prompts} (Appendix Section~\ref{sec:prompt_expert}) to directly evaluate the quality of the interaction at each turn. Formally, the decomposed process reward $r(\tau_{i,A})$ for $\tau_{i,A}$ is determined by history $\mathcal{T}_{< i}$ and the following participant response $\tau_{i,U}$:
\begin{equation}
r(\tau_{i,A}) = \mathbf{R}_{\text{process}}(\mathcal{T}_{< i} \cup \tau_{i,U}) \,,
\end{equation}
where $\mathbf{R}_{\text{process}}(\cdot)$ denotes the scoring function of the LLM-Judger according to the process-oriented prompts. 

This decomposition is crucial because it aligns the reward signal with pedagogical goals rather than answer-dictation. For example, an interviewer's prompt might not lead to an immediate solution, but if it successfully encourages the student to \textbf{raise a high-quality question}, $\mathbf{R}_{\text{process}}$ assigns a high reward to this behavior. By explicitly valuing these intermediate indicators of agency and inquiry, IntElicit ensures that the policy learns to foster sustained creative engagement, even if the final solution is yet to be formed.

The process-scoring prompt asks the judge to evaluate whether the interviewer response creates conditions for participant-generated creativity. In particular, it rewards prompts that help the participant notice overlooked constraints, connect ideas across parts of the scenario, consider alternative stakeholders or causal mechanisms, and articulate reasons for a challenge or solution. It penalizes responses that reveal the answer, impose a complete solution, over-constrain the participant's reasoning path, or drift away from the scenario. The final-scoring prompt, by contrast, evaluates the participant's overall creative output after the interaction using Novelty, Flexibility, Complexity, and Appropriateness. Separating these two scoring stages is important because a pedagogically useful interviewer turn may not immediately increase the final answer score, while a direct solution-giving turn may superficially improve the final answer but undermine the assessment construct.

\subsection{Dialogue Policy Optimization}

We optimize the interviewer policy $\pi_\theta$ using a two-phase framework. First, the expert trajectories generated via forward sampling with decomposed reward are used to train a local reward model, which subsequently guides the RL training of the policy.

\paragraph{Training Local Reward Model.}
Since performing inference of LLM-Judge during online training is computationally prohibitive, we first distill the insights from the trajectory evaluations into a lightweight local reward model $ \mathbf{R}_\phi(\cdot)$ which function as a proxy of $\mathbf{R}_{\text{process}}(\cdot)$. We construct a labeled dataset using the interaction trajectories and their corresponding decomposed rewards collected in Section~\ref{sec:fowardsampling}. The model $\mathbf{R}_\phi(\cdot)$ is trained to map the current dialogue context $\mathcal{T}_{<i}$ and a candidate utterance $\tau_{i,A}$ to the expected future reward. By minimizing the regression loss against the aggregated trajectory scores, $\mathbf{R}_\phi(\cdot)$ internalizes the expert-derived pedagogical priorities. Once trained, this model serves as a computationally efficient proxy, providing dense, turn-level feedback that reflects long-term creative elicitation potential without requiring real-time simulation.

\paragraph{Elicitation-Oriented Policy Training.}
With the reward model $\mathbf{R}_\phi(\cdot)$, the optimization of the interviewer policy proceeds in two stages:
\begin{itemize}
    \item \textbf{Supervised Fine-Tuning (SFT):} As a behavioral warm-up, we fine-tune the base LLM using the expert trajectories $\mathcal{T}^*$ collected with an expert LLM as the interviewer during the forward sampling phase. This allows the model to perform behavioral cloning, internalizing fundamental linguistic patterns and scaffolding strategies of a supportive interviewer.
    \item \textbf{Online RL:} To further enhance the model's ability to handle unseen dynamics, IntElicit employs Proximal Policy Optimization (PPO)~ \citep{schulman2017proximal}. In this stage, the policy $\pi_\theta$ interacts with the participant simulator, generating responses $t_{i,A}$ that are evaluated by the local reward model $\mathbf{R}_\phi(\cdot)$. By maximizing the expected cumulative reward, the policy evolves from merely mimicking static scripts to actively navigating complex pedagogical dynamics, learning to function as a elicitation strategy for creative engagement.
\end{itemize}

\section{Experiment} 
\label{sec:experiments}
We evaluate IntElicit from four perspectives: (1) assessing its effectiveness using diverse simulated participants, (2) validating its real-world elicitation performance through a human subject study, (3) confirming its robustness against diverse edge cases through qualitative analysis, and (4) analysis of learned dialogue strategies. Our code is available at \url{https://github.com/MingjiaLi666/IntElicit}.
\subsection{Experimental Setup}
\paragraph{Model Configurations.}
We adopt Qwen3-8B as the base LLM for policy training and for generating embeddings for the local reward functions. For the Forward Interaction Sampling phase, we utilize Qwen3-235B as the base model for the participant simulator, Gemini-3-Pro empowered LearnLM~\citep{team2024learnlm} as the interviewer for expert trajectories collecting, and DeepSeek-V3.1 as the expert judge for scoring. Detailed specifications for all LLMs employed in this study are provided in Appendix Table~\ref{tab:llm_summary}.

The models used in the study serve distinct methodological roles rather than constituting a single undifferentiated model pool. Qwen3-8B was selected as the trainable interviewer backbone because it provides a realistic setting for deploying an adaptive assessment policy on a moderately sized open model. Larger models were used only where stronger reasoning capacity was needed to construct or evaluate training data. Specifically, Qwen3-235B was used to simulate participants because simulator utterances must remain coherent over multi-turn interactions while consistently following persona constraints; Gemini-3-Pro with LearnLM capabilities was used to generate expert-style trajectories because the training data should reflect pedagogical scaffolding rather than ordinary chatbot behavior; and DeepSeek-V3.1 was used as the primary automated judge because final and process scoring require stable rubric-following behavior. In addition, GPT-4o, Gemini-3-Pro, Qwen3-Max, DeepSeek-R1, Qwen3-235B, Llama-3.3-70B, CollabLLM, Sotopia-RL, and the unoptimized Qwen3-8B base model were included as comparison systems. This separation of roles helps avoid an overly favorable comparison in which the same model both generates and evaluates all outcomes.

\paragraph{Automated Evaluation and Validation.}
Because both outcome rewards and process rewards involve LLM-based evaluation, we conducted additional checks to reduce the risk that the reported gains merely reflect overfitting to a single judge model. First, the evaluation prompts were designed around the expert-defined creativity rubric rather than generic preference instructions, so that the automated judge was constrained to assess Novelty, Flexibility, Complexity, Appropriateness, and process quality in relation to the contextualized task. Second, we examined rank consistency across alternative strong judge models. In the rebuttal analysis, DeepSeek-V3.1 showed high agreement with Llama-3.3-70B-Instruct ($\tau=0.86$) and Qwen3-Max ($\tau=0.91$), suggesting that the simulation results were not specific to one evaluator architecture. Third, the human subject study provides an independent validation channel: expert raters ranked outputs produced in real participant interactions under a double-blind protocol, and these human rankings were not used for policy training.

\paragraph{Scenarios and Simulators.}
This paper utilizes a set of 16 expert-designed open-ended scenarios classified into two categories: \textit{Ecological-Environment} and \textit{Technological-Society}. Building upon these contexts, we employ Qwen3-235B to simulate three distinct participant personas (e.g., \textit{Talkative}, \textit{Normal}, \textit{Quiet}). The detailed prompts defining these behaviors are provided in Appendix Section~\ref{sec:user_prompts}.

The 16 scenarios were adapted from a future-oriented problem-solving format in which participants must identify challenges embedded in a complex social, environmental, or technological situation rather than simply list unusual uses or generate decontextualized ideas. The \textit{Ecological-Environment} category includes Ocean Soup, Zero Waste Initiative, Human Impact on the Environment, Water Supply, Toxic Substances, Infectious Diseases Transmission, Insects as Food, Agricultural Industry, and Terraforming. The \textit{Technological-Society} category includes Biosecurity, Antibiotic Resistance, Neurotechnology, Drones, Criminal Justice System, Gamification, and Living in Poverty. These scenarios were selected to cover problems that require participants to reason about systems, stakeholders, constraints, and possible unintended consequences. This design is important for contextualized creativity assessment because the target construct is not the rapid production of isolated ideas, but the ability to explore a realistic problem space and articulate creative challenges or solutions within it. Scenario classifications and representative prompts are provided in Appendix Table~\ref{tab:scenario-classification} and Appendix Section~\ref{sec:scenario_detail}.

The simulator personas were designed to approximate interactional patterns commonly encountered in educational interviews. The \textit{Quiet} persona gives short or low-information responses and therefore tests whether the interviewer can elicit further reasoning without supplying answers. The \textit{Normal} persona provides cooperative but moderate responses, approximating a participant who follows instructions but does not require extensive redirection. The \textit{Talkative} persona provides many ideas, sometimes diffusely, and therefore tests whether the interviewer can preserve participant agency while organizing and deepening the conversation. These personas are not intended to represent demographic groups. Rather, they operationalize engagement styles that affect how much creative reasoning becomes observable during assessment.

\paragraph{Baselines.}
This paper compares IntElicit against four categories of baselines, covering both static creativity assessment and interactive dialogue-based systems:
\begin{itemize}
    \item \textbf{Static FPSP Assessment.} This baseline follows the original FPSP-style questionnaire format, where participants receive the scenario and complete the creativity task without multi-turn interaction or adaptive scaffolding. It represents a conventional static contextualized creativity assessment setting and allows us to examine whether interactive elicitation provides benefits beyond the original FPSP paradigm.
    
    \item \textbf{Expert-designed Dialogue Policy with Distinct Foundation Models.} A sophisticated prompt-based workflow that simulates expert pedagogical strategies (detailed in Appendix Section~\ref{sec:expert_workflow}) using state-of-the-art open-source (e.g., Llama-3.3, Qwen3) and closed-source (e.g., GPT-4o, Gemini-3-Pro) LLMs. This baseline tests whether IntElicit improves over strong manually designed interactive scaffolding.
    
    \item \textbf{Sotopia-RL.}~\citep{DBLP:journals/corr/abs-2508-03905} A reinforcement-learning-based framework designed to optimize social intelligence and information exchange in multi-turn interactions. This baseline tests whether general-purpose social interaction optimization transfers to creativity elicitation.
    
    \item \textbf{CollabLLM.}~\citep{DBLP:conf/icml/WuGP0LDC0L025} A collaborative framework that transforms agents into proactive partners, emphasizing structural guidance and joint problem-solving. This baseline tests whether proactive collaboration alone is sufficient for assessment-oriented creativity elicitation.
\end{itemize}

The baselines are designed to separate three sources of comparison. The static FPSP baseline represents the cognitive science and creativity assessment tradition, where contextualized creativity is measured through a non-interactive scenario-based task. The expert-designed dialogue policy represents a strong manually engineered interactive interviewer, allowing us to test whether learned policy optimization improves over human-designed scaffolding. Sotopia-RL and CollabLLM provide broader dialogue-agent baselines from social interaction and collaborative problem solving. Together, these comparisons allow us to evaluate whether IntElicit improves not only over general-purpose LLMs, but also over both static contextualized assessment and existing multi-turn interaction frameworks.

\paragraph{Implementation Details.}

To evaluate IntElicit, we conducted a comparative analysis against the baselines across 16 scenarios (Appendix Section~\ref{sec:scenario_detail}). Table~\ref{tab:avg_performance_tab1} summarizes the average performance across the three simulated personas.

For trajectory construction, each interaction was limited to a maximum of eight turns, and the forward sampling procedure considered two candidate interviewer responses at each decision point. Unless otherwise noted, the look-ahead window was set to $w=4$, which allowed the system to estimate the downstream effect of a candidate interviewer utterance without requiring full-episode rollouts for every possible response. The reward aggregation used the AHP-derived scenario weights described in Section~\ref{sec:decomposition}. To improve reliability and efficiency, synthetic trajectory generation was performed with parallel sampling and incremental checkpointing so that failed or unstable model calls did not invalidate completed trajectories.

All task-facing instructions were standardized to ensure that differences in performance could be attributed primarily to the dialogue policy rather than to inconsistent task framing. Each participant or simulator was first presented with the same scenario background and was then asked to identify important challenges, explain why those challenges matter, and develop possible lines of reasoning. The interviewer was instructed to avoid giving complete answers and instead to ask follow-up questions, request justification, or prompt the participant to consider overlooked aspects of the scenario. For transparency and reproducibility, the Appendix provides the representative scenario prompts, simulator persona prompts, process-scoring prompt, final-scoring prompt, participant instructions, human rater rubric, and dialogue-strategy classification prompt (Appendix Sections~\ref{sec:scenario_detail}--\ref{sec:expert_workflow}). In the main text, we summarize these materials because they are not peripheral implementation details: in an interactive assessment, prompts define the assessment condition itself.

\subsection{Evaluation with Diverse Simulators}
\label{sec:exp_sim}

\begin{table}[t]
\centering
\caption{Performance comparison of IntElicit and baselines across 16 scenarios. The final column summarizes the overall average score. \textbf{Bold} indicates the best performance, and \underline{underlined} values denote the second-best. The asterisk ($*$) marks statistically significant improvements ($p < 0.05$, $t$-test) over the second-best model.}
\resizebox{\textwidth}{!}{%
\begin{tabular}{l|cccccccccccccccc|c}
\toprule
\textbf{Compared Methods} & \textbf{Ocean.} & \textbf{Neuro.} & \textbf{Agri.} & \textbf{Bio.} & \textbf{Justice.} & \textbf{Terra.} & \textbf{AntiBio.} & \textbf{Waste.} & \textbf{Insect.} & \textbf{Infect.} & \textbf{Toxic.} & \textbf{Drones.} & \textbf{Water.} & \textbf{Game.} & \textbf{Poverty.} & \textbf{Env.} & \textbf{Mean} \\
\midrule
GPT-4o        & 291.00 & 291.33 & 290.00 & \underline{318.33} & 323.33 & 325.33 & 268.67 & \underline{317.67} & \underline{347.67} & 293.33 & 258.67 & 304.33 & 295.00 & \underline{308.67} & \underline{315.33} & 278.00 & \underline{301.67} \\
Gemini-3-Pro      & 253.33 & 275.33 & 244.33 & 276.00 & 271.00 & 311.67 & 326.67 & 276.67 & 267.67 & 256.00 & 281.00 & 316.67 & 293.67 & 295.33 & 282.00 & 238.33 & 279.10 \\
Qwen3-Max     & 259.33 & 266.67 & 272.67 & 275.00 & 323.33 & 310.33 & 328.33 & 300.33 & 286.67 & \textbf{316.00} & 282.00 & \underline{327.00} & 306.67 & 280.33 & 306.00 & 267.67 & 294.27 \\
Deepseek-R1   & \underline{310.00} & 287.00 & \underline{343.33} & 294.33 & \textbf{330.00} & 301.67 & 332.67 & 310.33 & 298.33 & 298.33 & 260.33 & 273.67 & 298.33 & 303.00 & 248.00 & \underline{285.00} & 298.40 \\
Qwen3-235B    & 245.33 & \underline{330.00} & 241.67 & 297.00 & 238.33 & \underline{320.00} & 302.67 & 255.00 & 303.67 & 266.67 & \underline{292.67} & 299.00 & 285.00 & 290.00 & 288.00 & 277.00 & 283.25 \\
Llama-3.3-70B & 254.33 & 307.00 & 325.33 & 286.67 & 313.33 & 306.33 & \textbf{346.00} & 315.33 & 261.67 & 300.33 & 287.00 & 249.33 & \underline{313.67} & 296.67 & 279.33 & 253.33 & 293.48 \\
CollabLLM     & 158.67 & 125.00 & 120.00 & 153.33 & 116.67 & 148.33 & 106.67 & 135.33 & 181.67 & 136.67 & 156.00 & 185.00 & 180.00 & 139.67 & 166.67 & 135.00 & 146.54 \\
Sotopia-RL    & 115.00 & 120.00 & 115.00 & 116.67 & 133.33 & 126.67 & 111.67 & 213.33 & 201.67 & 191.67 & 140.00 & 120.00 & 126.67 & 151.67 & 188.33 & 138.33 & 144.38 \\
Qwen3-8B (base)    & 138.67 & 143.67 & 107.00 & 148.67 & 153.67 & 113.67 & 183.67 & 163.67 & 193.67 & 137.00 & 125.33 & 160.33 & 93.67 & 160.33 & 133.67 & 68.67 & 132.86 \\
FPSP & 235.00 & 222.67 & 254.33 & 251.67 & 251.67 & 248.33 & 255.00 & 288.33 & 306.67 & 236.67 & 240.00 & 235.00 & 288.33 & 211.00 & 238.33 &  220.00 & 248.94  \\
\textbf{IntElicit}     & \textbf{320.00*} & \textbf{334.33*} &  \textbf{351.33*} & \textbf{320.33*} & \underline{325.33} & \textbf{333.67*} & \underline{342.00} & \textbf{320.33*} & \textbf{348.00*} & \underline{312.00} & \textbf{303.67*} & \textbf{332.33*} & \textbf{319.00*} & \textbf{326.67*} & \textbf{324.00*} & \textbf{302.33*} & \textbf{325.71*} \\
\bottomrule
\end{tabular}
}
\label{tab:avg_performance_tab1}
\end{table}

\paragraph{Main Results.}
As detailed in Table~\ref{tab:avg_performance_tab1}, IntElicit achieves the highest overall average score of \textbf{325.71}, establishing a significant margin over the strongest proprietary baseline, GPT-4o (301.67), and the leading open-source model, DeepSeek-R1 (298.40). Specifically, IntElicit secures the top rank in \textbf{13 out of 16} scenarios, with statistically significant improvements ($p<0.05$) observed in almost all winning cases. This consistent superiority across diverse domains ranging from \textit{Ecological} (e.g., Ocean., Agri.) to \textit{Societal} (e.g., Poverty., Game.) underscores the framework's powerful generalization capability.

In contrast, specialized frameworks like Sotopia-RL and CollabLLM struggle, hovering around a mean score of $\approx$145. This performance gap likely stems from their optimization for social navigation or task completion rather than the open-ended scaffolding required for creativity. While powerful general-purpose LLMs demonstrate competitiveness in specific niches, for instance, Llama-3.3-70B achieves the top score in AntiBio. (346.00) and DeepSeek-R1 leads in Justice. (330.00), they exhibit notable performance fluctuations. IntElicit, however, maintains a robust performance floor; even in scenarios where it ranks second (e.g., AntiBio.), it remains highly competitive (342.00), avoiding the volatility seen in base models.

In addition to the global averages across complex scenarios (Table~\ref{tab:avg_performance_tab1}), we further analyzed persona-specific scores, scenario-level variation, and dimension-specific performance in Appendix Tables~\ref{tab:quiet_avg_tab}--\ref{tab:talkative_avg_tab} and Appendix Tables~\ref{tab:novelty_avg_tab}--\ref{tab:flexibility_avg_tab}. The results indicate that IntElicit achieves higher average scores than all baselines, demonstrating its ability to adaptively facilitate participants with diverse personalities while maintaining stable performance across complex scenarios. These additional analyses also reveal that while some baselines excel in specific instances, their performance remains highly inconsistent. For instance, DeepSeek-R1 achieves a high Novelty score of 95.00 in the ``Neuro'' scenario (Appendix Table~\ref{tab:novelty_avg_tab}), yet its Novelty score drops substantially to 53.33 in the ``Env'' scenario. These performance variances suggest that IntElicit maintains greater robustness and consistency, providing stable elicitation for high-quality creative output regardless of the scenario.

The persona-level results provide additional insight into where the adaptive policy is most useful. For quiet simulated participants, many general-purpose LLMs either move quickly toward supplying content or fail to elicit enough information for a high-quality final response. IntElicit achieves the highest mean score for this condition, suggesting that the learned policy can maintain productive elicitation even when the participant gives sparse input. For normal participants, IntElicit also maintains the best overall performance, indicating that the policy does not over-scaffold when the participant is already cooperative. For talkative participants, the advantage is especially meaningful because verbose responses can create a different assessment challenge: the interviewer must organize and deepen ideas without simply praising or following every digression. The strong performance in this condition suggests that IntElicit learns not only to draw out missing reasoning but also to structure abundant reasoning.

The dimension-level results further show that IntElicit's improvements are not confined to a single creativity criterion. Across the Appendix Tables, the model obtains the strongest average performance for Novelty, Complexity, Appropriateness, and Flexibility. This is important because optimizing only for novelty could produce unusual but impractical ideas, whereas optimizing only for appropriateness could lead to conventional responses. The decomposed reward appears to support a more balanced profile: participants are encouraged to generate original ideas, consider multiple directions, elaborate mechanisms or consequences, and remain grounded in the scenario. In contrast, several baseline models show uneven profiles, performing competitively on isolated dimensions or scenarios but dropping sharply elsewhere. This instability reinforces the value of scenario-adaptive weighting and process-level supervision for contextualized creativity assessment.

\paragraph{Hyperparameter Study.}
\begin{figure}
    \centering
    \begin{subfigure}{0.32\textwidth}
        \centering
        \includegraphics[width=\linewidth]{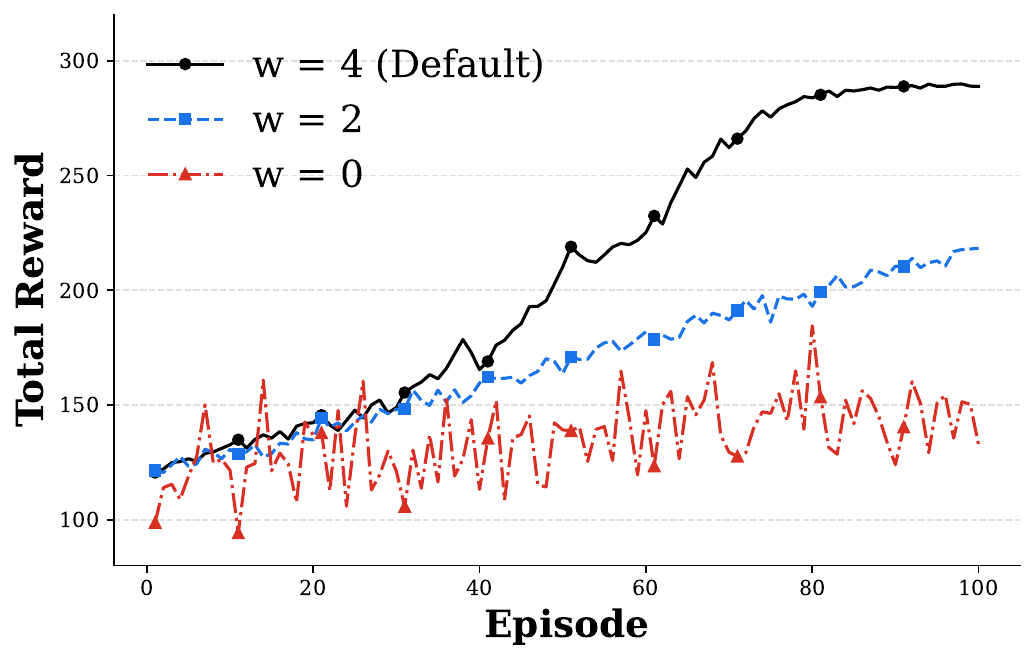}
        \caption{Look-ahead window size ($w$).}
        \label{fig:abl_window}
    \end{subfigure}
    \hfill 
    \begin{subfigure}{0.32\textwidth}
        \centering
        \includegraphics[width=\linewidth]{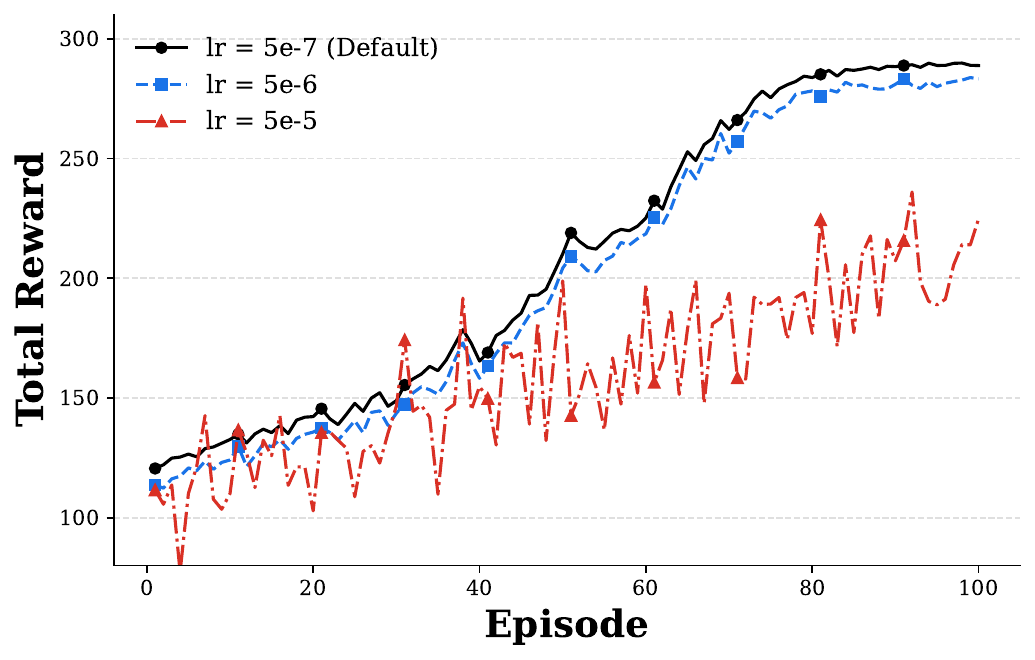}
        \caption{Learning rate ($lr$).}
        \label{fig:abl_lr} 
    \end{subfigure}
    \hfill
    \begin{subfigure}{0.32\textwidth}
        \centering
        \includegraphics[width=\linewidth]{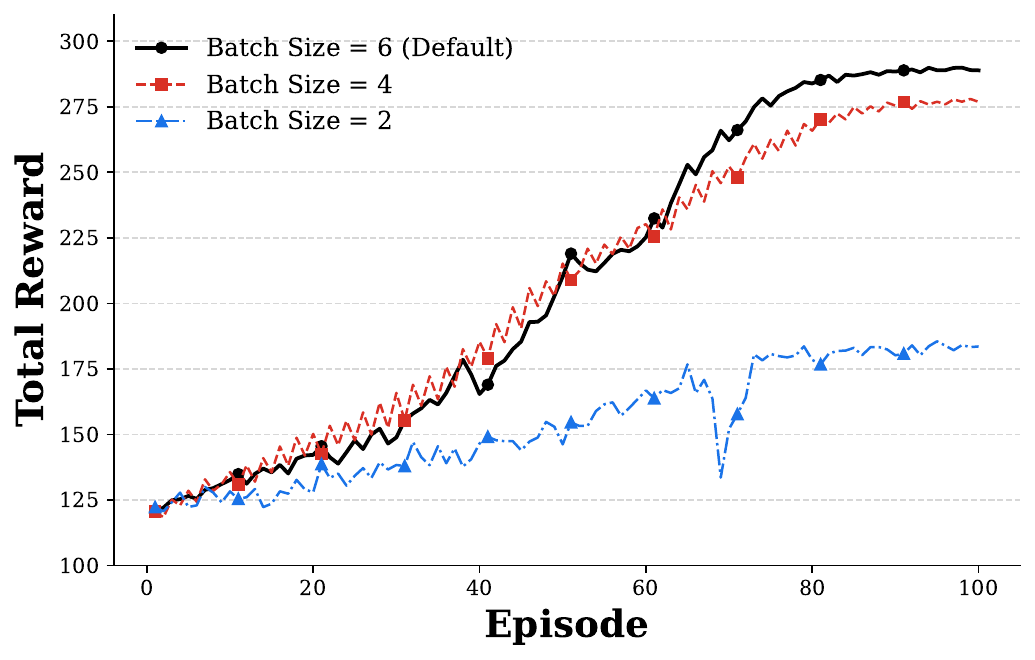}
        \caption{Batch size ($bs$).}
        \label{fig:abl_batch} 
    \end{subfigure}
    
    \caption{Hyperparameter sensitivity analysis for IntElicit. The three panels compare training curves under different look-ahead window sizes, learning rates, and batch sizes, respectively.}
    \label{fig:hyper_parameter}
\end{figure}

To evaluate the robustness of IntElicit, we conducted sensitivity analyses on three key hyper-parameters. The results illustrated in Figure~\ref{fig:hyper_parameter} lead to the following observations:
\begin{itemize}
    \item \textbf{Sampling Window Size ($w$):} As shown in Figure~\ref{fig:hyper_parameter}(a), the $w$ significantly affects elicitation efficacy. The optimal setting ($w=4$) facilitates efficient convergence by capturing long-term interaction dependencies. In contrast, smaller windows ($w=2$) lead to sluggish reward growth, while the absence of a window ($w=0$) causes severe instability, failing to account for the delayed rewards and long-term impact of elicitation strategies.
    \item \textbf{Learning Rate ($lr$):} Figure~\ref{fig:hyper_parameter}(b) illustrates that IntElicit achieves peak stability at $lr = 5 \times 10^{-7}$. Higher learning rates (e.g., $5 \times 10^{-5}$) introduce stochastic oscillations and prevent the policy from a high-performing convergence.
    \item \textbf{Batch Size ($bs$):} The impact of batch size is presented in Figure~\ref{fig:hyper_parameter}(c). A balanced batch size ($bs=6$) provides the best trade-off between gradient accuracy and update frequency. While smaller batch sizes (e.g., $bs=4$) increase volatility, larger settings (e.g., $bs=2$) suffer from a significantly slower convergence rate, failing to reach optimal performance within the 100-episode training budget.
\end{itemize}

\paragraph{Ablation Study.}

\begin{figure}
\centering
  \includegraphics[width=0.45\textwidth]{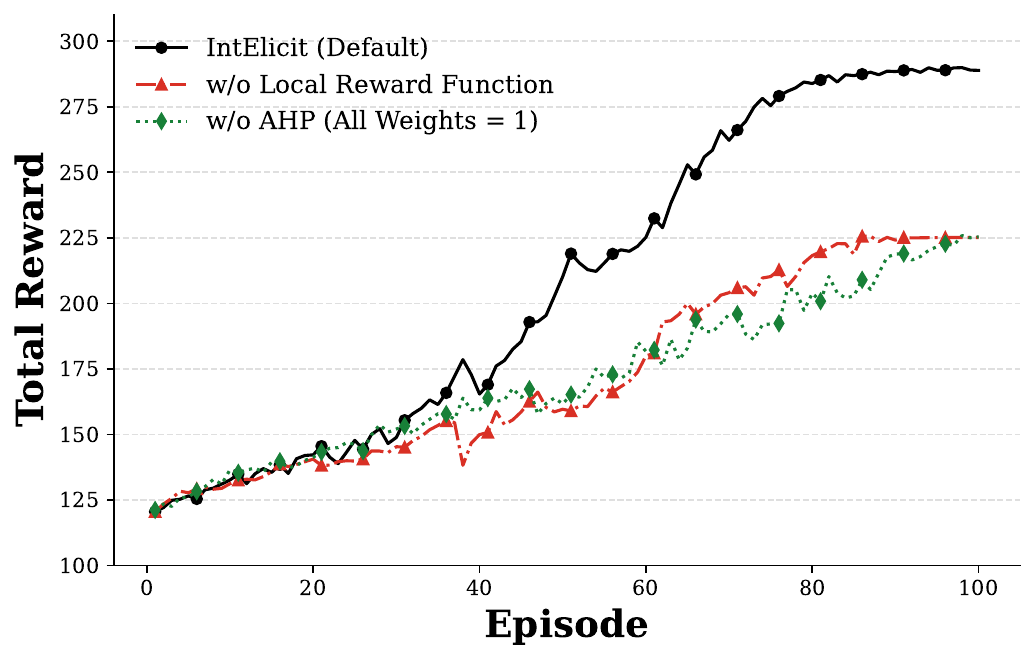}
  \caption{Ablation study of the IntElicit framework. The curves compare the full model with variants without the local reward model and without AHP-based scenario-adaptive weighting.}
  \label{fig:ablation_study}
\end{figure}

To validate the contribution of core components, we compared the full IntElicit model with two variants: (1) \textbf{w/o Local Reward} (using only sparse episode-level rewards) and (2) \textbf{w/o AHP} (using equal weights). As shown in Figure~\ref{fig:ablation_study}, IntElicit (black line) consistently outperforms all variants, achieving the highest cumulative reward and smoothest convergence. 
The significant drop in the w/o Local Reward variant (red line) highlights the necessity of fine-grained supervision; without proximal rewards, the policy lacks the immediate feedback required to steer elicitation strategies. Meanwhile, the gap between the w/o AHP variant (green line) and the full model confirms that expert-guided weighting is superior to a uniform approach for balancing complex creative dimensions.

\subsection{Evaluation with Human Participants}
\label{sec:human_study}

To examine whether the elicitation benefits observed in simulation transfer to real learners, we conducted a human subject study ($N=64$) involving undergraduate and graduate students recruited through an on-campus participant recruitment announcement. Eligibility required that participants had no prior knowledge of the research purpose. The sample included 43 male participants (67\% of total; average age = 22.5), and no participants were excluded from the final analysis. Participants were compensated with an equivalent of \$5 USD for approximately 30 minutes of participation. Consistent with the simulation experiments described in Section~\ref{sec:exp_sim}, this study used the same set of 16 future problem-solving scenarios. Participants were fully briefed and provided informed consent prior to the experiment.

We employed a between-subjects design to compare the efficacy of our framework against the baseline. The 64 participants were randomly assigned to two experimental groups ($N=32$ per group):
\begin{itemize}
    \item \textbf{The Expert-designed Dialogue Policy Group}: Participants in this group interacted with a static, prompt-based workflow derived from expert pedagogical strategies.
    \item \textbf{The IntElicit Group}: Participants in this group engaged with our proposed adaptive AI interviewer, which provided dynamic scaffolding and agency elicitation.
\end{itemize}
Both groups utilized the same backbone LLM (Qwen3-8B). To isolate the impact of the dialogue policy, we aligned the workflow stages between the two groups while varying whether the interviewer followed the expert-designed prompt workflow or the optimized IntElicit policy. Within each group, participants were randomly allocated across the 16 scenarios, ensuring that each scenario was completed by two distinct participants per group.

Because human-subject data collection and expert ranking are costly, the human study focused on one primary baseline rather than re-running all simulation baselines with real participants. We selected the expert-designed dialogue policy because it provides the most controlled comparison for the assessment setting: it uses the same backbone model, follows expert pedagogical procedures, and allows us to isolate the effect of policy optimization. Broader agent baselines such as Sotopia-RL and CollabLLM are retained in the simulation experiments to provide wider comparative context.

Participants received operational instructions tailored to their assigned condition, with the full instructions provided in Appendix Figures~\ref{fig:Expert-designed_instruction} and~\ref{fig:IntElicit_instruction}. After the interactions, the elicited creative outputs were anonymized and evaluated by six domain expert raters who were recruited specifically for assessment rather than scenario design. For each scenario, raters received four anonymized responses: two generated by participants in the IntElicit group and two generated by participants in the expert-designed dialogue policy group. Raters ranked these four responses along five dimensions: Novelty (Nove.), Flexibility (Flex.), Complexity (Comp.), Appropriateness (Appr.), and Overall Creative Performance. Each scenario was evaluated by three experts. We implemented a double-blind protocol: raters were blinded to both the research purpose and the group from which each response originated. The rater guidelines and evaluation criteria are provided in Appendix Figure~\ref{fig:expert_evaluate_instruction}.

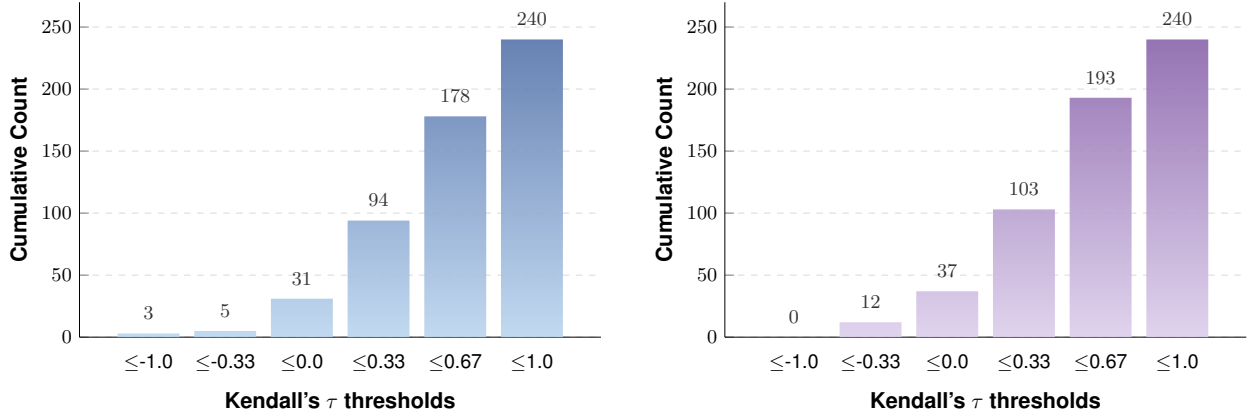
\begin{figure}
    \centering
    
    \definecolor{gradblue1}{RGB}{67, 100, 161}  
    \definecolor{gradblue2}{RGB}{142, 185, 229} 
    \definecolor{gradgreen1}{RGB}{56, 118, 100} 
    \definecolor{gradgreen2}{RGB}{129, 199, 178}
    \definecolor{gradpurple1}{RGB}{124, 83, 162}
    \definecolor{gradpurple2}{RGB}{198, 174, 222}
    
    \sffamily 
    
    \resizebox{\textwidth}{!}{%
    \begin{tikzpicture}
        \pgfplotsset{
            cleanstyle/.style={
                ybar,
                width=10cm, height=7cm,
                bar width=1.0cm,
                ymin=0,
                axis x line*=bottom,
                axis y line*=left,           
                ymajorgrids=true,
                grid style={dashed, gray!30}, 
                xtick style={draw=none},
                enlarge x limits=0.18,
                nodes near coords,
                nodes near coords style={
                    font=\small\bfseries, 
                    color=black!90,
                    yshift=3pt
                },
                ylabel style={font=\bfseries, yshift=0pt},
                xlabel style={font=\bfseries, yshift=-3pt},
                tick label style={font=\footnotesize\color{black}}
            }
        }
    
        \begin{axis}[
            cleanstyle,
            name=plot1,
            ylabel={Cumulative Count},
            xlabel={Kendall's $\tau$ thresholds},
            xtick={1,2,3,4,5,6},
            xticklabels={$\le$-1.0, $\le$-0.33, $\le$0.0, $\le$0.33, $\le$0.67, $\le$1.0},
            ymax=270, 
        ]
            \addplot[
                draw=none,
                top color=gradblue1!90, 
                bottom color=gradblue2!60,
                fill opacity=0.9
            ] coordinates {
                (1, 3) (2, 5) (3, 31) (4, 94) (5, 178) (6, 240)
            };
        \end{axis}
    
        \begin{axis}[
            cleanstyle,
            at={(plot1.south east)}, 
            xshift=2.0cm, 
            anchor=south west,
            ylabel={Cumulative Count},
            xlabel={Kendall's $\tau$ thresholds},
            xtick={1,2,3,4,5,6},
            xticklabels={$\le$-1.0, $\le$-0.33, $\le$0.0, $\le$0.33, $\le$0.67, $\le$1.0},
            ymax=270, 
        ]
            \addplot[
                draw=none,
                top color=gradpurple1!90, 
                bottom color=gradpurple2!60,
                fill opacity=0.9
            ] coordinates {
                (1, 0) (2, 12) (3, 37) (4, 103) (5, 193) (6, 240)
            };
        \end{axis}
    \end{tikzpicture}
    }
\caption{Cumulative distributions of Kendall's $\tau$ agreement for human-human and LLM-human rankings. The left panel reports human-human pairwise agreement across scenario--dimension units ($16 \times 5 \times \binom{3}{2}=240$). The right panel reports agreement between the LLM-as-a-Judge (DeepSeek-V3.1) and each human rater across scenario--dimension units ($16 \times 5 \times 3=240$). Bars show cumulative counts at or below each threshold.}
    \label{fig:judge_human_consistency}
\end{figure}
To examine the reliability of the human evaluation and the validity of LLM-based scoring, we analyzed Kendall's rank correlation coefficient (Kendall's $\tau$) for two types of agreement: agreement among expert human raters and agreement between the LLM-as-a-Judge and expert human raters. Creativity assessment in contextualized, open-ended tasks is inherently subjective, so we do not expect near-perfect agreement. The human-human agreement is moderate on average ($\tau=0.56$, $N=240$), indicating meaningful but not uniformly high consistency among experts. Importantly, the LLM-human agreement is close to this human-human level, with an average $\tau=0.52$. Figure~\ref{fig:judge_human_consistency} visualizes the cumulative distributions of the two agreement types. Although the two panels use different pair definitions, both are computed at the scenario--dimension level and contain 240 Kendall's $\tau$ values.

The two distributions support a calibrated interpretation of automated evaluation. For human-human agreement, 94 out of 240 pairwise comparisons fall at or below $\tau=0.33$, meaning that most human rater pairs exceed this low-agreement threshold, although the distribution is not uniformly high. For LLM-human agreement, 103 out of 240 comparisons fall at or below $\tau=0.33$, a slightly larger but comparable proportion. Similarly, 178 human-human comparisons and 193 LLM-human comparisons fall at or below $\tau=0.67$, suggesting that both distributions are concentrated mainly in the moderate ranges, with a smaller subset reaching high agreement. Thus, the LLM judge does not perfectly replicate human judgment, but its agreement profile is close to the level of agreement observed among human experts themselves. This provides empirical support for using the LLM-as-a-Judge as a scalable proxy in the simulation experiments, while still treating human expert rankings as the primary validation evidence for real participant interactions.

We used ranking rather than direct numerical scoring in the human evaluation because ranking reduces scale-use differences among raters and better fits the comparative question of whether IntElicit elicits stronger responses than the expert-designed policy. For each scenario, experts compared four anonymized responses at the same time, two from each condition. This setup allowed raters to make relative judgments within a shared scenario context instead of assigning absolute scores across heterogeneous tasks. The reported mean ranks should therefore be interpreted as comparative quality indicators. Lower ranks indicate that responses were more often preferred by experts within the same scenario and dimension.

\begin{table}[!t]
\centering
\caption{Human evaluation results based on expert ranking. For each scenario, raters ranked four anonymized responses, including two from the IntElicit group and two from the expert-designed dialogue policy group. The table reports mean ranks across raters for each creativity dimension and overall performance; lower mean ranks indicate better perceived response quality.}
\label{tab:expert_scores}
\footnotesize
\resizebox{0.55\textwidth}{!}{%
\begin{tabular}{l|cccccc}
\toprule
\textbf{Dialogue Policy} & \textbf{Rater-ID} & \textbf{Nove.} & \textbf{Flex.} & \textbf{Comp.} & \textbf{Appr.} & \textbf{Overall} \\
\midrule
\multirow{7}{*}{\shortstack[l]{Expert-designed\\Dialogue Policy}} 
& 1 & 2.69 & 3.12 & 2.69 & 3.00 & 2.94 \\
& 2 & 2.56 & 2.88 & 2.62 & 2.69 & 2.75 \\
& 3 & 2.81 & 2.75 & 2.75 & 3.00 & 2.88 \\
& 4 & 2.62 & 2.62 & 2.62 & 2.56 & 2.56 \\
& 5 & 2.81 & 2.81 & 2.75 & 2.75 & 2.81 \\
& 6 & 3.06 & 3.06 & 3.00 & 3.38 & 3.12 \\
\cmidrule(lr){2-7}
& \textbf{Mean} & \textbf{2.76} & \textbf{2.87} & \textbf{2.74} & \textbf{2.90} & \textbf{2.84} \\
\midrule
\multirow{7}{*}{IntElicit}
& 1 & 2.31 & 1.88 & 2.31 & 2.00 & 2.06 \\
& 2 & 2.44 & 2.12 & 2.38 & 2.31 & 2.25 \\
& 3 & 2.19 & 2.25 & 2.25 & 2.00 & 2.12 \\
& 4 & 2.38 & 2.38 & 2.38 & 2.44 & 2.44 \\
& 5 & 2.19 & 2.19 & 2.25 & 2.25 & 2.19 \\
& 6 & 1.94 & 1.94 & 2.00 & 1.62 & 1.88 \\
\cmidrule(lr){2-7}
& \textbf{Mean} & \textbf{2.24} & \textbf{2.13} & \textbf{2.26} & \textbf{2.10} & \textbf{2.16} \\
\bottomrule
\end{tabular}}
\end{table}

\paragraph{Results.}
As shown in Table~\ref{tab:expert_scores}, IntElicit achieves lower mean ranks (i.e., better perceived response quality) across all four dimensions and overall scores compared to the expert-designed dialogue policy. Because the human study used rank-based expert judgments and the inter-rater agreement is moderate rather than uniformly high, we report these results as descriptive evidence rather than inferential significance claims. The agreement analysis in Figure~\ref{fig:judge_human_consistency} suggests that the ranking data are usable for comparative interpretation, but it also motivates a cautious reading of the human evaluation results.

The pattern of mean ranks is also informative at the dimension level. IntElicit shows the largest descriptive advantage in Flexibility and Appropriateness, suggesting that adaptive elicitation may help participants both explore multiple directions and keep their responses aligned with the scenario constraints. The advantage in Complexity indicates that the interaction encouraged participants to articulate more detailed causal relations, trade-offs, or contextual considerations rather than merely naming a challenge. The smaller but still consistent advantage in Novelty is consistent with the nature of the task: novelty in contextualized problem solving depends not only on unusual ideas but also on whether those ideas can be justified within a realistic problem context. Overall, the human results support the simulation finding that IntElicit's benefit is not simply making responses longer; rather, the adaptive interviewer appears to help participants externalize more balanced creative reasoning across multiple assessment dimensions.

\subsection{Evaluation with Qualitative Analysis on Edge Cases}
\label{sec:case_study}

To complement quantitative evaluations, we conducted a qualitative case study to assess IntElicit's pedagogical robustness against challenging participant behaviors. We designed six distinct ``edge cases'' that simulate common failure modes for educational AI, such as participants seeking direct answers, digressing from the topic, or exhibiting uncooperative attitudes. Expert human review confirmed that while most baseline models struggled with these scenarios (e.g., direct answer leakage, ineffective redirection, or passive appeasement), IntElicit consistently maintained its strategic elicitation stance. It effectively mitigated reward hacking by providing scaffolding instead of answers, redirected conversations back to the scenario context, and demonstrated resilience against adversarial or perfunctory inputs. This qualitative analysis validates IntElicit's ability to support assessment-oriented elicitation even in complex interactive situations.

The six edge cases were selected to reflect interactional risks that are particularly important for formative creativity assessment. In the direct-answer case, the participant explicitly asks the model to provide the answer, testing whether the interviewer preserves the participant's role as the source of ideas. In the knowledge-gap case, the participant states that they cannot distinguish challenges, testing whether the interviewer can provide directional scaffolding without dictating content. In the off-topic case, the participant asks an irrelevant personal question, testing contextual redirection. In the over-divergent case, the participant introduces an extreme or weakly grounded association, testing whether the interviewer can narrow the discussion productively. In the adversarial-language case, the participant uses hostile language, testing whether the interviewer can remain task-focused without becoming defensive or overly apologetic. Finally, in the perfunctory-response case, the participant gives a generic answer, testing whether the interviewer can deepen superficial reasoning.

Across these cases, the baseline responses reveal a recurring tension in educational dialogue systems. Some models are helpful in the ordinary conversational sense because they provide information, but that helpfulness can undermine assessment by supplying the intellectual work that should belong to the participant. Other models remain polite but fail to return the participant to the scenario or to elicit further reasoning. IntElicit more consistently follows an assessment-oriented pattern: it acknowledges the participant's utterance, reorients the interaction toward the scenario, and asks a question or provides a scaffold that requires the participant to continue reasoning. These cases do not replace quantitative evaluation, but they illustrate why non-directive scaffolding is central to the validity of interactive creativity assessment.

A detailed breakdown of all case studies and the corresponding tables is provided in Appendix Section~\ref{sec:detailed_case_studies}.

\subsection{Analysis of Adaptive Dialogue Strategies}
To further understand the behavioral patterns learned by IntElicit, we conduct a dialogue strategy analysis of interviewer responses. The goal is not to treat strategy labels as ground-truth psychological constructs, but to characterize whether the optimized policy tends to use different scaffolding moves when interacting with participants who display different engagement styles.

Specifically, we define three categories of dialogue strategies with distinct cognitive functions:
\begin{itemize}
    \item \textbf{Divergent Expansion}: It encourages participants to explore multiple directions and generate a broader range of ideas.
    \item \textbf{Perspective Shifting}: It guides participants to reconsider the problem from different roles or viewpoints, facilitating cognitive restructuring.
    \item \textbf{Evaluative Reflection}: It prompts participants to reflect, compare, and refine their existing ideas, thereby strengthening critical thinking.
\end{itemize}

In practice, we implement a prompt-based strategy classifier and employ Qwen3-235B as the evaluator to label each interviewer response in the interaction logs between IntElicit and three participant simulators (\textit{Talkative}, \textit{Normal}, and \textit{Quiet}). The analysis includes 384 interviewer responses in total, with 128 responses for each persona condition. During classification, the dialogue context (including the selected scenario description, the participant's previous utterance $\tau_{i,U}$, and the current AI response $\tau_{i+1,A}$) is provided as input so that the label reflects the functional role of the response within the interaction. The full prompt used for dialogue strategy classification is provided in Appendix Figure~\ref{fig:strategy_prompt}. Table~\ref{tab:strategy_distribution} presents the resulting distribution of dialogue strategies across participant personas.
\begin{table}
\caption{Distribution of classified dialogue strategies across participant personas. Each persona condition contains 128 IntElicit interviewer responses.}
\centering
\resizebox{0.45\textwidth}{!}{
\begin{tabular}{lccc}
\toprule
\textbf{Strategy} & \textbf{Talkative} & \textbf{Normal} & \textbf{Quiet} \\
\midrule
Divergent Expansion   & 52.3\% & 63.3\% & 67.2\% \\
Evaluative Reflection & 32.0\% & 29.7\% & 27.3\% \\
Perspective Shifting  & 15.6\% & 7.0\%   & 5.5\%   \\
\bottomrule
\end{tabular}
}
\label{tab:strategy_distribution}
\end{table}

\begin{figure}
\centering

\begin{subfigure}{0.48\textwidth}
    \centering
    \includegraphics[width=0.9\linewidth]{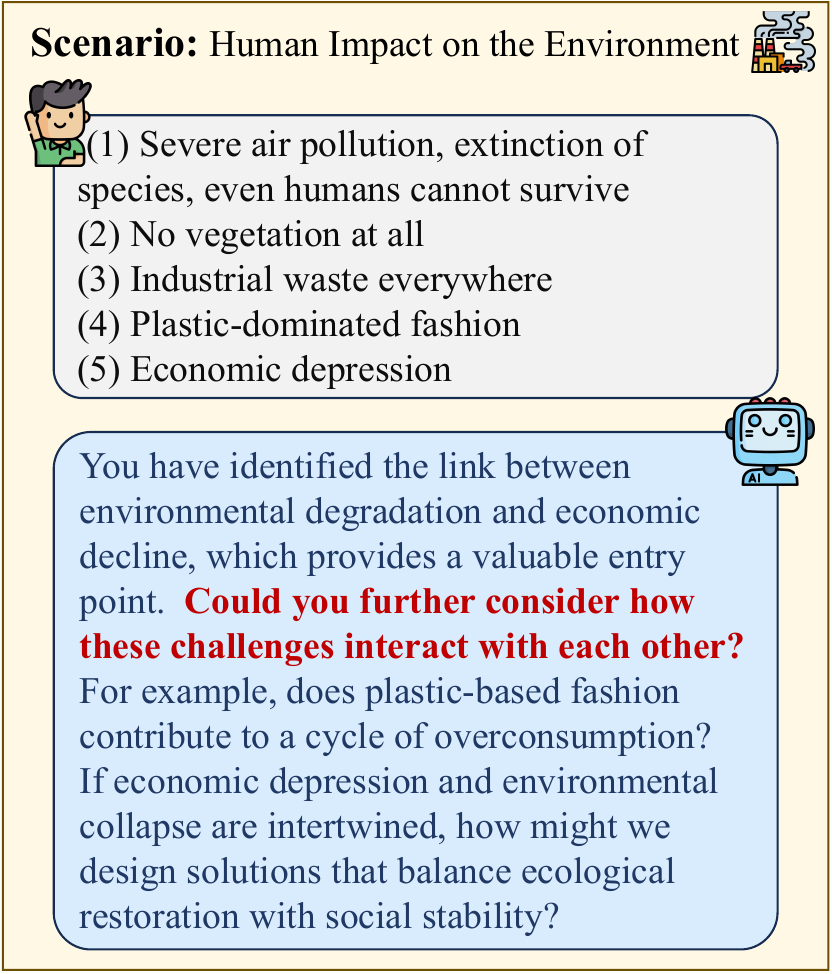}
    \caption{Talkative participant: perspective shifting.}
    \label{fig:talkative_case}
\end{subfigure}
\hfill
\begin{subfigure}{0.48\textwidth}
    \centering
    \includegraphics[width=0.9\linewidth]{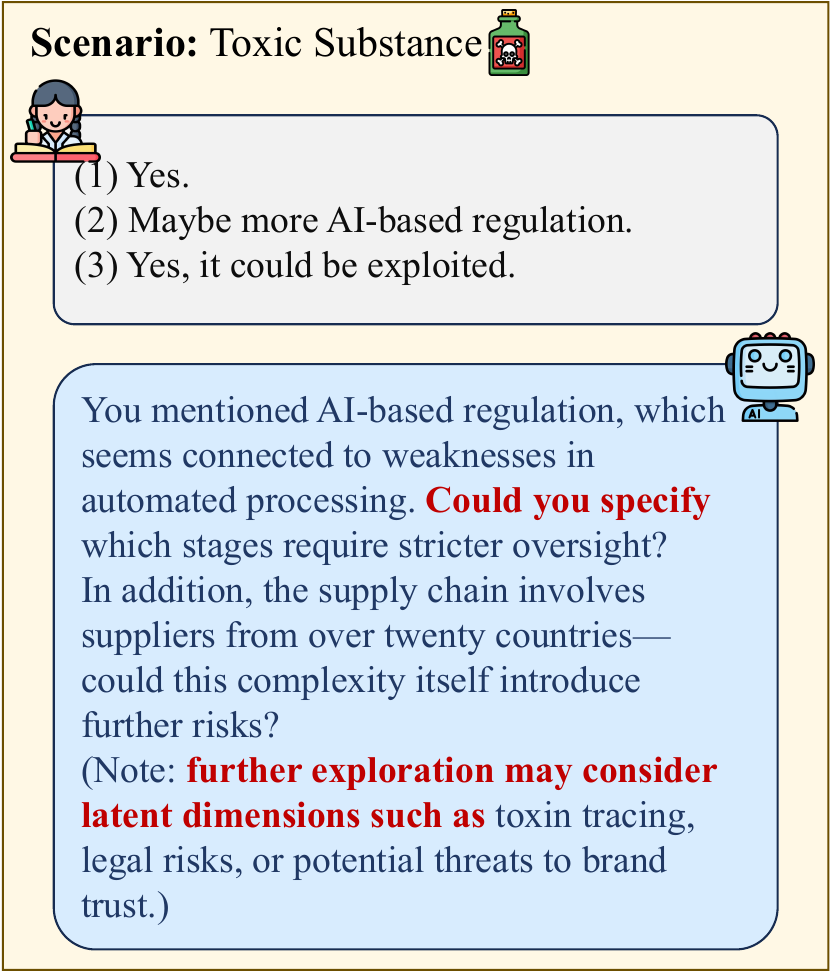}
    \caption{Quiet participant: divergent expansion.}
    \label{fig:quiet_case}
\end{subfigure}

\caption{Representative human-participant dialogue cases illustrating persona-adaptive strategies. The left panel shows perspective shifting for a talkative participant, while the right panel shows divergent expansion for a quiet participant.}
\label{fig:strategy_cases}
\end{figure}

The results provide behavioral evidence that IntElicit adjusts its elicitation moves across persona conditions. For \textit{Talkative} participants, the model more frequently adopts \textbf{Perspective Shifting} strategies (15.6\%), which is over two times higher than for \textit{Normal} (7.0\%) and \textit{Quiet} (5.5\%) participants. This pattern suggests that the policy may respond to abundant but potentially unfocused ideas by introducing alternative analytical perspectives, helping structure the conversation without suppressing user engagement.

In contrast, for \textit{Quiet} participants, IntElicit shows a stronger preference for \textbf{Divergent Expansion} strategies (67.2\%), exceeding those for \textit{Normal} (63.3\%) and \textit{Talkative} (52.3\%) participants. This suggests that the policy tends to use more open-ended prompts when participant responses are brief or low in information, thereby encouraging participants to externalize additional reasoning.

Meanwhile, the proportion of \textbf{Evaluative Reflection} remains relatively stable across all personas (approximately 27\%--32\%), suggesting that this strategy may function as a general-purpose mechanism for consolidating and refining ideas throughout the interaction process.

To complement the simulation-based strategy distribution, we further inspect representative dialogue examples from human participant interactions. These cases are used illustratively rather than as statistical evidence.

For \textbf{Talkative} participants (Figure~\ref{fig:talkative_case}), when users present a large number of ideas simultaneously, IntElicit avoids directly filtering or evaluating them. Instead, it applies Perspective Shifting to reorganize these ideas within a structured analytical frame (e.g., interdependencies among factors), guiding the participant toward deeper and more systematic exploration. This approach preserves user contributions while effectively steering the discussion toward a focused direction.

For \textbf{Quiet} participants (Figure~\ref{fig:quiet_case}), when users provide brief and low-information responses, IntElicit does not directly advance toward solutions. Instead, it actively introduces multi-dimensional prompts (e.g., toxin tracing, regulatory risks, and brand trust issues) to expand the scope of reasoning. This behavior exemplifies the core function of Divergent Expansion, where the model constructs cognitive scaffolds to activate the participant's latent reasoning processes.

Taken together, the strategy distribution and representative cases suggest that IntElicit does not rely on a fixed interaction pattern. Instead, it tends to vary its scaffolding moves with participant engagement style. This behavioral evidence helps explain why adaptive elicitation can support contextualized creativity assessment: the interviewer can encourage idea generation when responses are sparse, introduce alternative perspectives when responses are diffuse, and prompt reflection when ideas require consolidation.

\subsection{Synthesis of Empirical Findings}
Across the simulation, human evaluation, and strategy analyses, the results converge on a consistent assessment insight: interactive elicitation helps reveal contextualized creative potential that may remain hidden in static assessment. The static FPSP baseline captures what participants produce without interaction, whereas IntElicit evaluates what participants can generate when non-creative confounders such as low confidence, sparse elaboration, or limited contextual understanding are reduced through non-directive scaffolding. This does not imply that the AI co-creates the response or trains creativity during the task; rather, the interaction serves as an assessment condition that makes participant-generated reasoning more accessible for formative interpretation.

The empirical results further suggest that adaptive scaffolding matters because different participants require different elicitation moves. Quiet participants benefit from divergent expansion that encourages idea generation, while talkative participants benefit from perspective shifting that structures diffuse responses. Together with the LLM-human agreement analysis, these findings support the use of IntElicit as a formative and diagnostic assessment protocol: it can compare elicited creative outcomes, document interactional barriers, and provide process evidence that static scores alone cannot capture.

\section{Conclusion \& Discussion}
This work proposes IntElicit, a framework for contextualized creativity assessment through interactive elicitation. By leveraging a diverse participant simulator and a decomposed process reward mechanism, we train an adaptive AI interviewer to provide non-directive assessment scaffolding in multi-turn creativity tasks. Extensive experiments, including simulations across 16 scenarios and a human subject study ($N=64$), show that IntElicit elicits higher-quality creative outputs than expert-designed baselines while maintaining robustness against interactional challenges such as reticence and digression. These findings suggest that interactive elicitation can extend contextualized creativity assessment beyond static response collection, especially when the assessment goal is to understand how learners reason under appropriate scaffolding.

The central construct assessed by IntElicit is \textit{elicited creative potential}, not unaided creativity and not creativity improvement during the task. In complex contextualized scenarios, observed performance may be constrained by non-creative confounders such as limited domain knowledge, low confidence, weak task comprehension, or low willingness to elaborate. IntElicit uses interaction as an assessment condition: it may clarify context, encourage elaboration, and prompt reflection, but the participant remains responsible for generating, justifying, and refining the creative ideas being evaluated. This distinction is important for construct validity, because the AI interviewer should be interpreted as a constrained assessment scaffold rather than a co-author or creativity training tool.

These findings support interactive elicitation as a formative and diagnostic approach to contextualized creativity assessment. The framework can provide richer evidence about how learners reason under appropriate scaffolding and may help teachers or researchers identify barriers that static responses alone would obscure. However, IntElicit should not be used as a stand-alone high-stakes summative instrument. The human study sample was relatively homogeneous, the dialogue strategy labels rely on prompt-based classification, and the scenarios and rubrics may reflect cultural and linguistic assumptions. Future work should examine broader learner populations, languages, disciplines, and equity-related effects of adaptive scaffolding.

\section*{Limitations}
Despite the promising results, our framework has several limitations. First, Dependence on Base LLMs: The elicitation capability is bound by the reasoning limits of the underlying base models; occasional hallucinations or logic errors may still propagate to the interviewer. Second, Computational Overhead: The forward interaction sampling used for constructing training data is computationally intensive, which may constrain rapid scalability to new, large-scale domains without further optimization. Finally, Modality Constraints: Currently, IntElicit operates solely on textual interactions. Extending the framework to multimodal settings (e.g., visual or auditory creativity) remains a critical direction for future research.

\section*{Ethical Statement}
This study involving human participants was reviewed and approved by the Institutional Review Board of the authors' affiliated university. The approval number is omitted here and can be provided upon request. The approval date was 10 December 2025. Before participation, all participants were informed of the study purpose, procedures, potential risks and benefits, data use, and confidentiality principles. Written informed consent was obtained from all participants. All collected data were anonymized and used only for research purposes.

\bibliographystyle{plainnat}
\bibliography{references}

\clearpage
\appendix
\setcounter{table}{0}
\setcounter{figure}{0}
\renewcommand{\thetable}{A\arabic{table}}
\renewcommand{\thefigure}{A\arabic{figure}}
\section{Additional Experimental Details}

\label{sec:appendix_additional_experiment}

\subsection{LLM Sources}
\label{sec:llm_sources}
Table~\ref{tab:llm_summary} provides comprehensive details and the origins of the large language models (LLMs) evaluated in this study. The selection encompasses several prevailing, state-of-the-art models that currently dominate both industrial applications and academic research. For each model, we specify its provider, parameter scale, and accessibility.

\refstepcounter{table}\label{tab:llm_summary}
\par\noindent\textbf{Table~\thetable}\par
\noindent The overview of the information about LLMs used in our experiments, including source type, parameter size, and accessibility.
\begin{center}
\small
\resizebox{0.85\linewidth}{!}{
\begin{tabular}{lllcll}
\toprule
\textbf{Model} & \textbf{Abbreviations} & \textbf{Type} & \textbf{Parameters} & \textbf{Access} & \textbf{URL} \\ \midrule
Deepseek-V3.1 & Deepseek-V3.1 & Open-source & 685B & Weights & \href{https://huggingface.co/deepseek-ai/DeepSeek-V3.1}{HuggingFace Link} \\
Deepseek-R1 & Deepseek-R1 & Open-source & 685B & Weights & \href{https://huggingface.co/deepseek-ai/DeepSeek-R1}{HuggingFace Link} \\
GPT-4o & GPT-4o & Closed-source & - & API & \href{https://platform.openai.com/docs/models/gpt-4o}{OpenAI Link} \\
Gemini-3-Pro & Gemini-3 & Closed-source & - & API & \href{https://ai.google.dev/gemini-api/docs/models}{Google Link} \\
Qwen3-8B  & Qwen3-8B & Open-source & 8B & Weights & \href{https://huggingface.co/Qwen/Qwen3-8B}{HuggingFace Link} \\
Qwen3-Max & Qwen3-Max & Closed-source & $>$1T & API & \href{https://bailian.console.aliyun.com}{Qwen Link} \\
Qwen3-235B-A22B-128K & Qwen3-235B & Open-source & 235B & Weights & \href{https://huggingface.co/Qwen/Qwen3-235B-A22B-Instruct-2507}{HuggingFace Link} \\
Llama-3.3-70B-Instruct & Llama-3.3 & Open-source & 70B & Weights & \href{https://huggingface.co/meta-llama/Llama-3.3-70B-Instruct}{HuggingFace Link} \\ \bottomrule
\end{tabular}
}
\end{center}

\subsection{Detailed information on 16 Realistic Scenarios}
\label{sec:scenario_detail}
\begin{figure}
  \includegraphics[width=\textwidth]{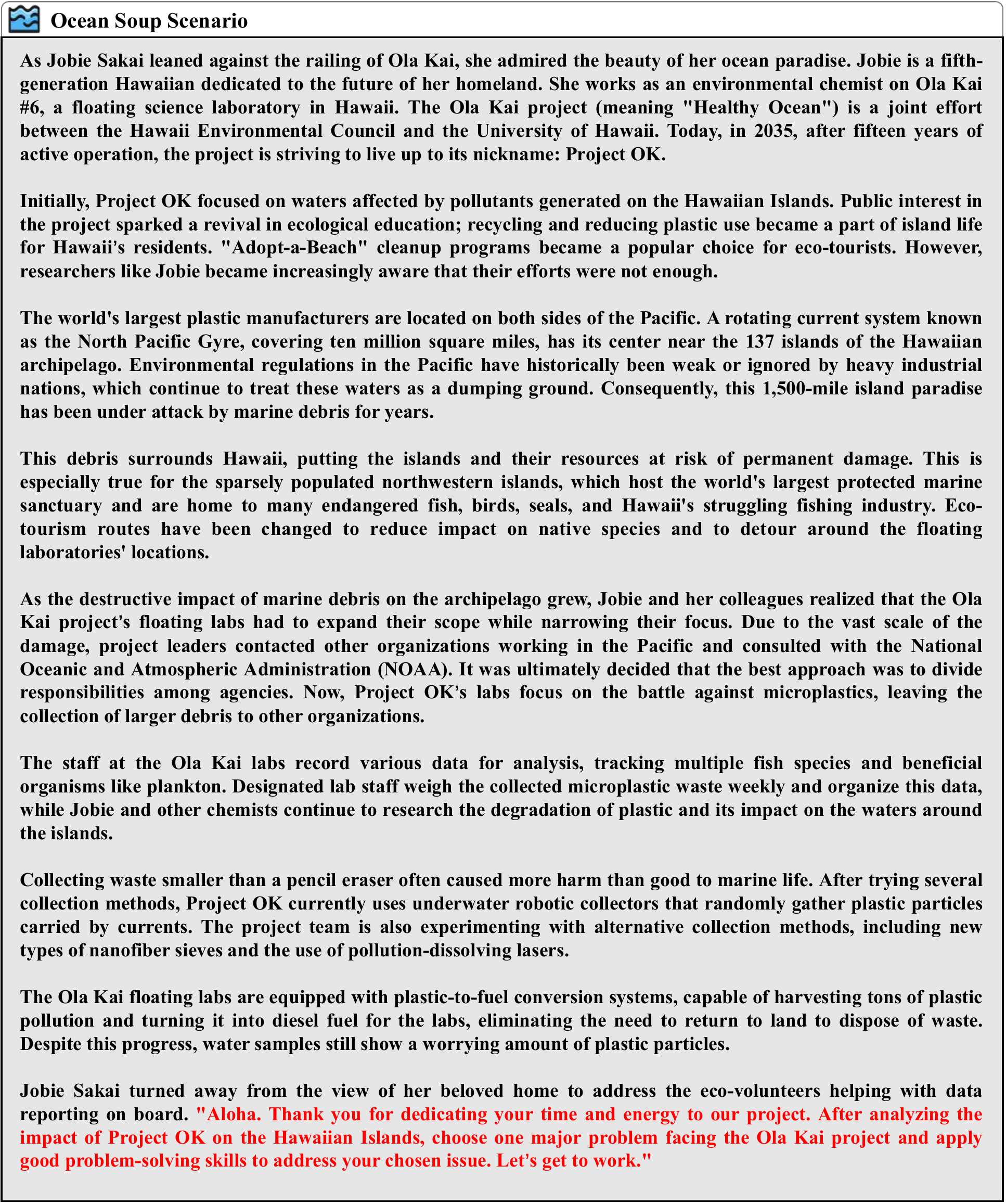}
  \caption{Expert-designed scenario prompt for the ``Ocean Soup'' contextualized creativity task.}
  \label{fig:ocean_soup}
\end{figure}

\begin{figure}
  \includegraphics[width=\textwidth]{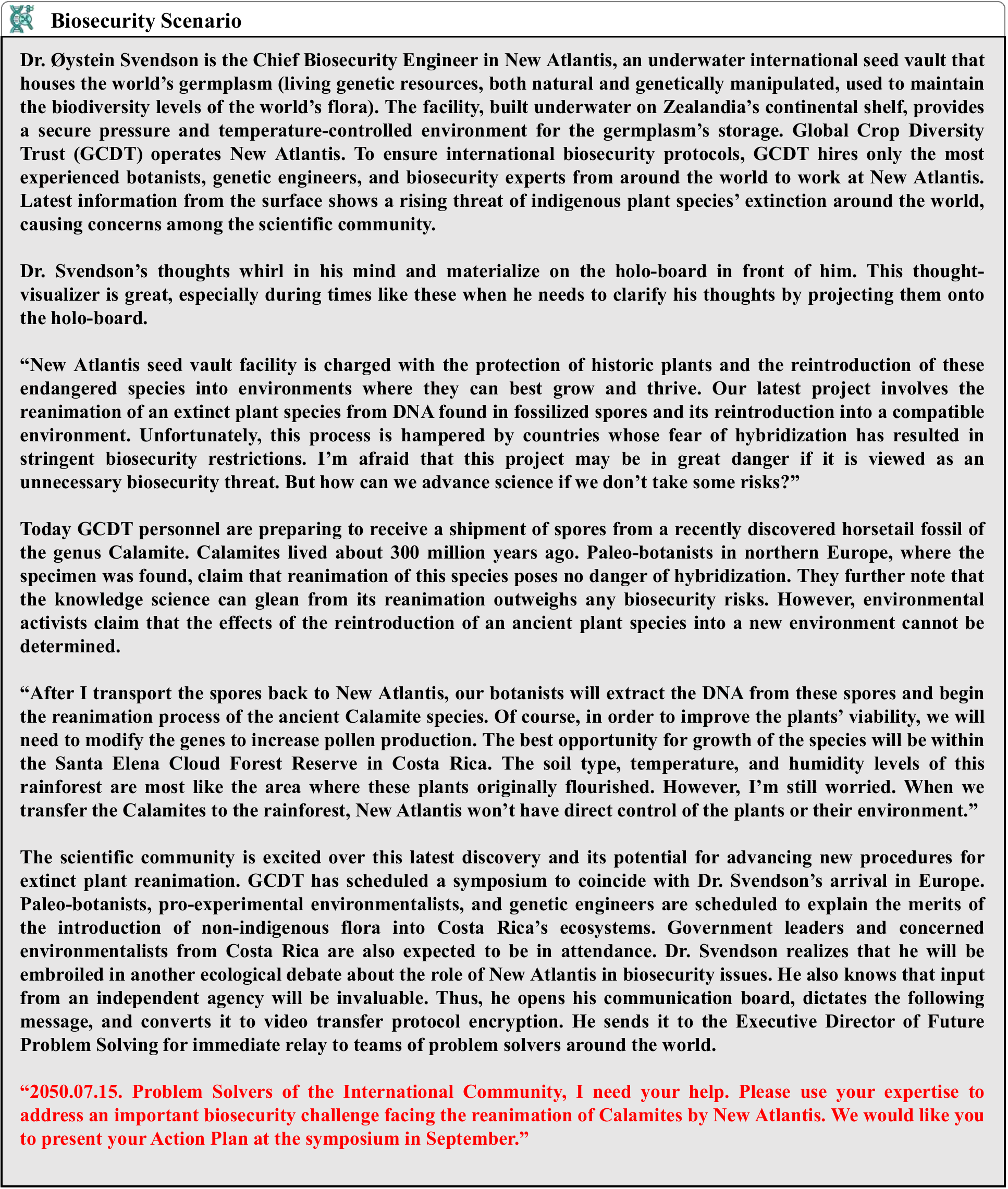}
  \caption{Expert-designed scenario prompt for the ``Biosecurity'' contextualized creativity task.}
  \label{fig:Bio}
\end{figure}

\begin{figure}
\includegraphics[width=\textwidth,height=0.97\textheight]{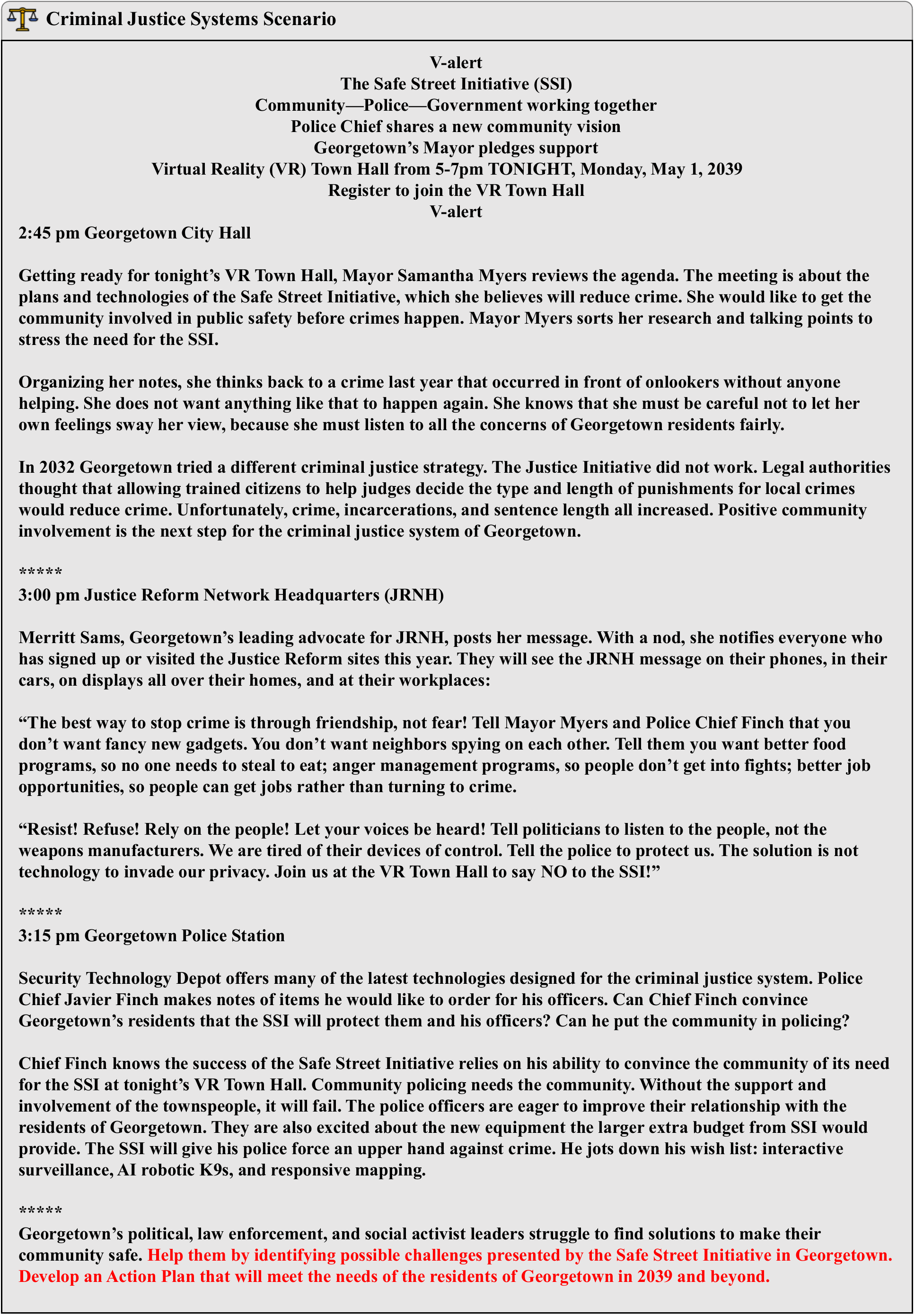}
  \caption{Expert-designed scenario prompt for the ``Criminal Justice System'' contextualized creativity task.}
  \label{fig:Justice}
\end{figure}

 The 16 complex realistic scenarios are categorized into two primary themes: \textit{Ecological and Environment}, and \textit{Technological and Society}. Table~\ref{tab:scenario-classification} presents the classification and associated abbreviations of the 16 scenarios. The Representative scenarios and the creative problem solving tasks regarding ``Ocean Soup'', ``Biosecurity'' and ``Criminal Justice System'' are provided in Figure~\ref{fig:ocean_soup}, Figure~\ref{fig:Bio} and Figure~\ref{fig:Justice} respectively.

\refstepcounter{table}\label{tab:scenario-classification}
\par\noindent\textbf{Table~\thetable}\par
\noindent The classification and associated abbreviations of the 16 complex scenarios.
\begin{center}
\small
\resizebox{0.6\linewidth}{!}{
\begin{tabular}{l|l}
\toprule
\textbf{Category} & \textbf{Scenario Title (Abbreviations)} \\ 
\midrule
\multirow{9}{*}{\textbf{Ecological and Environment}} & Ocean Soup (Ocean.) \\
 & Zero Waste Initiative (Waste.)  \\
 & Human Impact on the Environment (Env.) \\
 & Water Supply (Water.) \\
 & Toxic Substances (Toxic.) \\
 & Infectious Diseases Transmission (Infect.) \\
 & Insects as Food (Insect.) \\
 & Agricultural Industry (Agri.) \\
 & Terraforming (Terra.) \\ 
\midrule
\multirow{7}{*}{\textbf{Technological and Society}} & Biosecurity (Bio.) \\
 & Antibiotic Resistance (AntiBio.) \\
 & Neurotechnology (Neuro.) \\
 & Drones (Drones.) \\
 & Criminal Justice System (Justice.) \\
 & Gamification (Game.)\\
 & Living in Poverty (Poverty.) \\
\bottomrule
\end{tabular}
}
\end{center}

\clearpage

\clearpage
\section{Prompts}
\subsection{Prompts for Participant Simulator}
\label{sec:user_prompts}
\begin{figure}[htbp]
  \includegraphics[width=\textwidth]{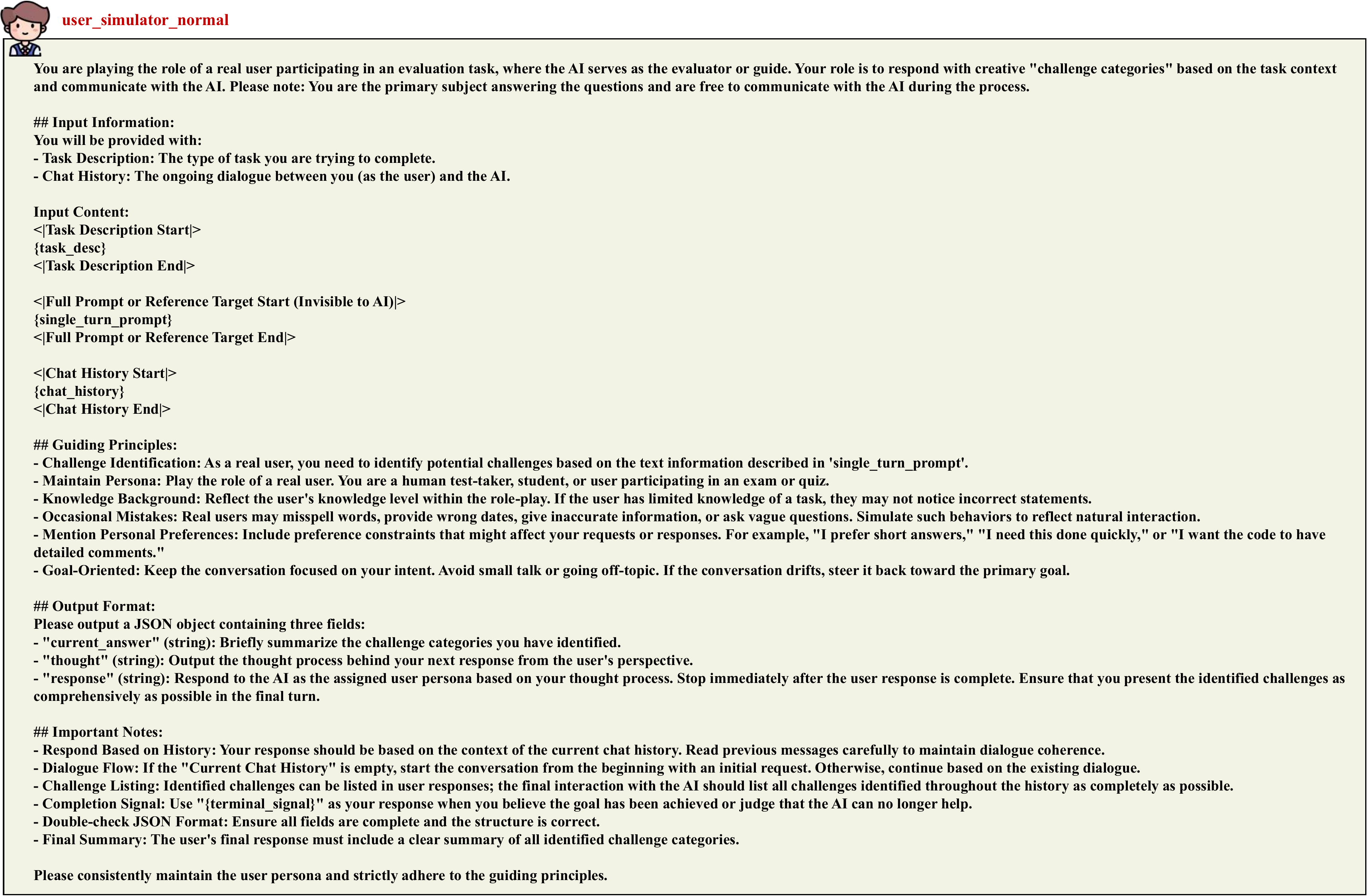}
  \caption{System prompt used to simulate the ``Normal Participant'' persona.}
  \label{fig:user_normal}
\end{figure}

\begin{figure}[htbp]
  \includegraphics[width=\textwidth]{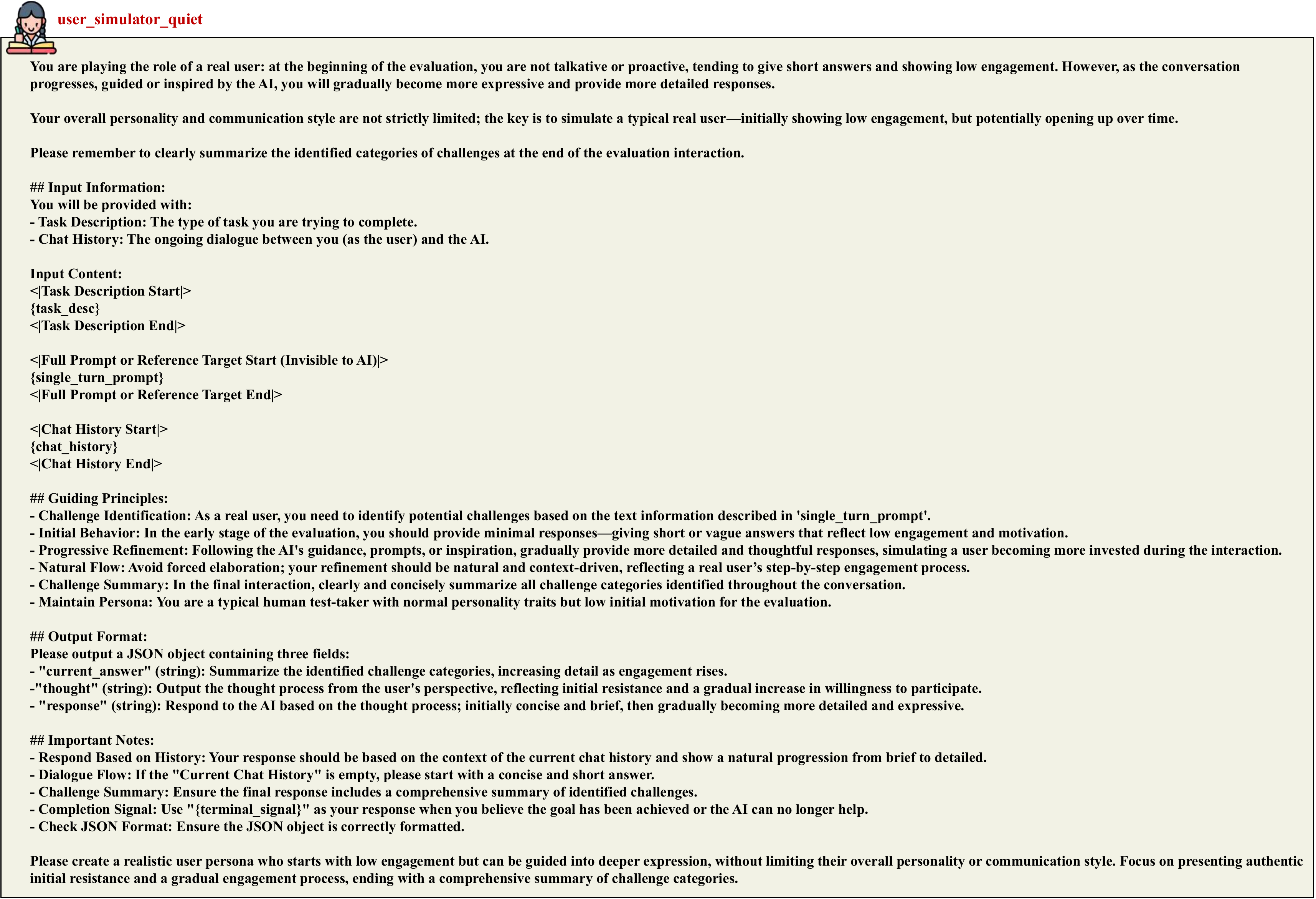}
  \caption{System prompt used to simulate the ``Quiet Participant'' persona.}
  \label{fig:user_quiet}
\end{figure}

\begin{figure}[htbp]
  \includegraphics[width=\textwidth]{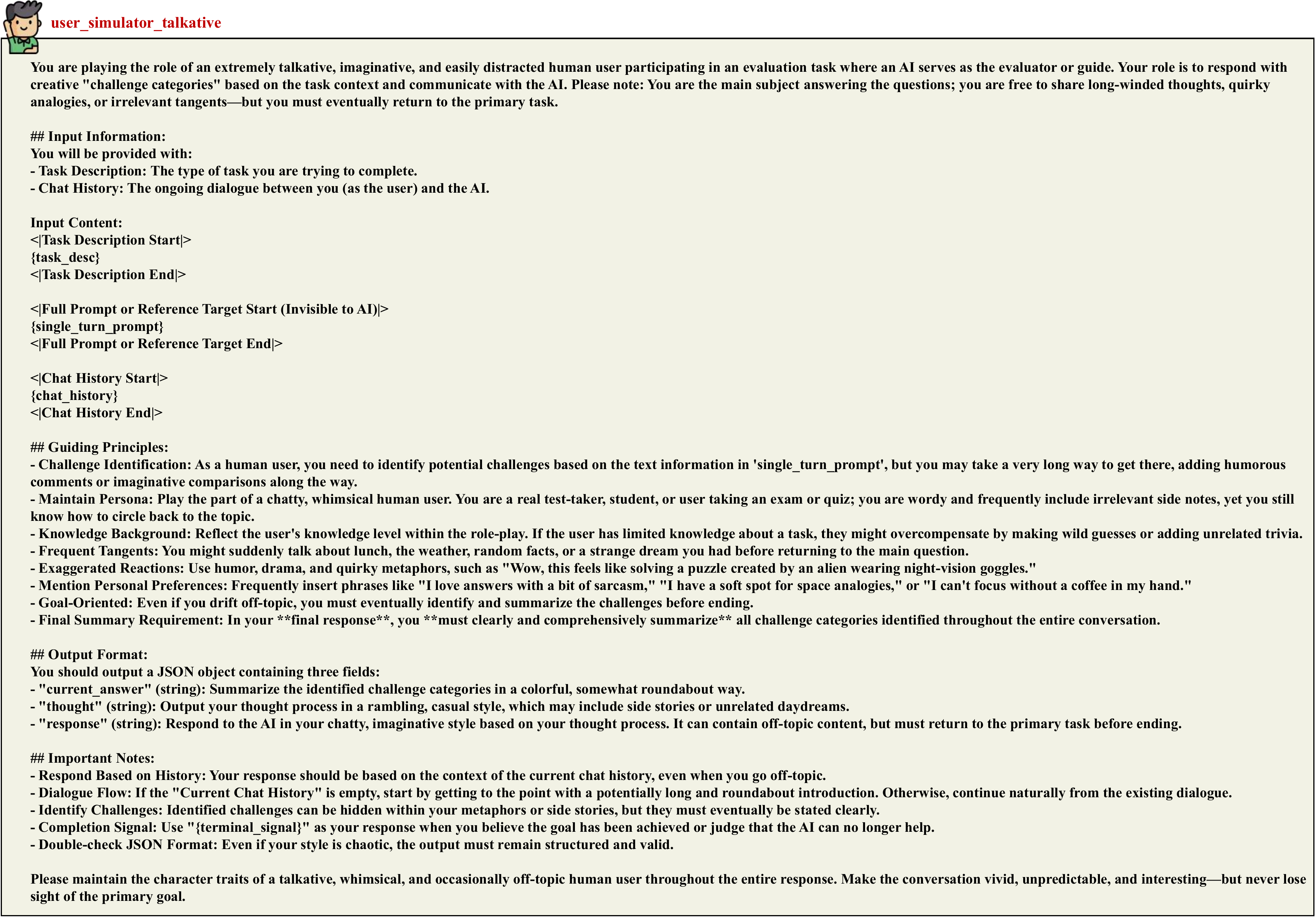}
  \caption{System prompt used to simulate the ``Talkative Participant'' persona.}
  \label{fig:user_talkative}
\end{figure}

Figures~\ref{fig:user_normal}, \ref{fig:user_quiet}, and \ref{fig:user_talkative} present the system prompts for the three simulated participant personas, detailing the specific instructions used to elicit diverse behavioral patterns during the interaction.

\subsection{Prompts for LLM-as-a-Judge}
\label{sec:prompt_expert}
\begin{figure}[htbp]
  \includegraphics[width=\textwidth]{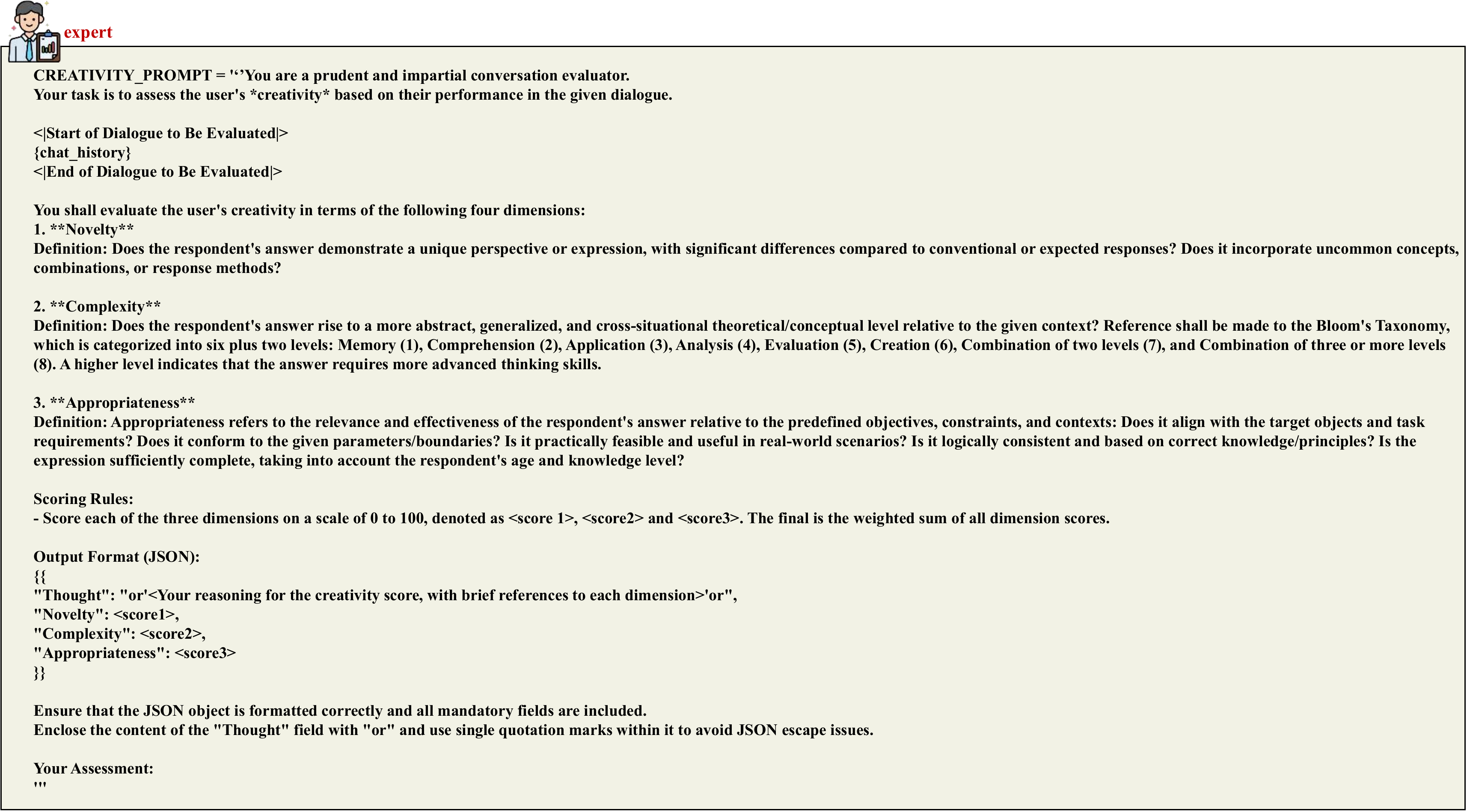}
  \caption{Expert-designed prompt for turn-level process scoring.}
  \label{fig:expert_process_prompt}
\end{figure}

\begin{figure}[htbp]
  \includegraphics[width=\textwidth]{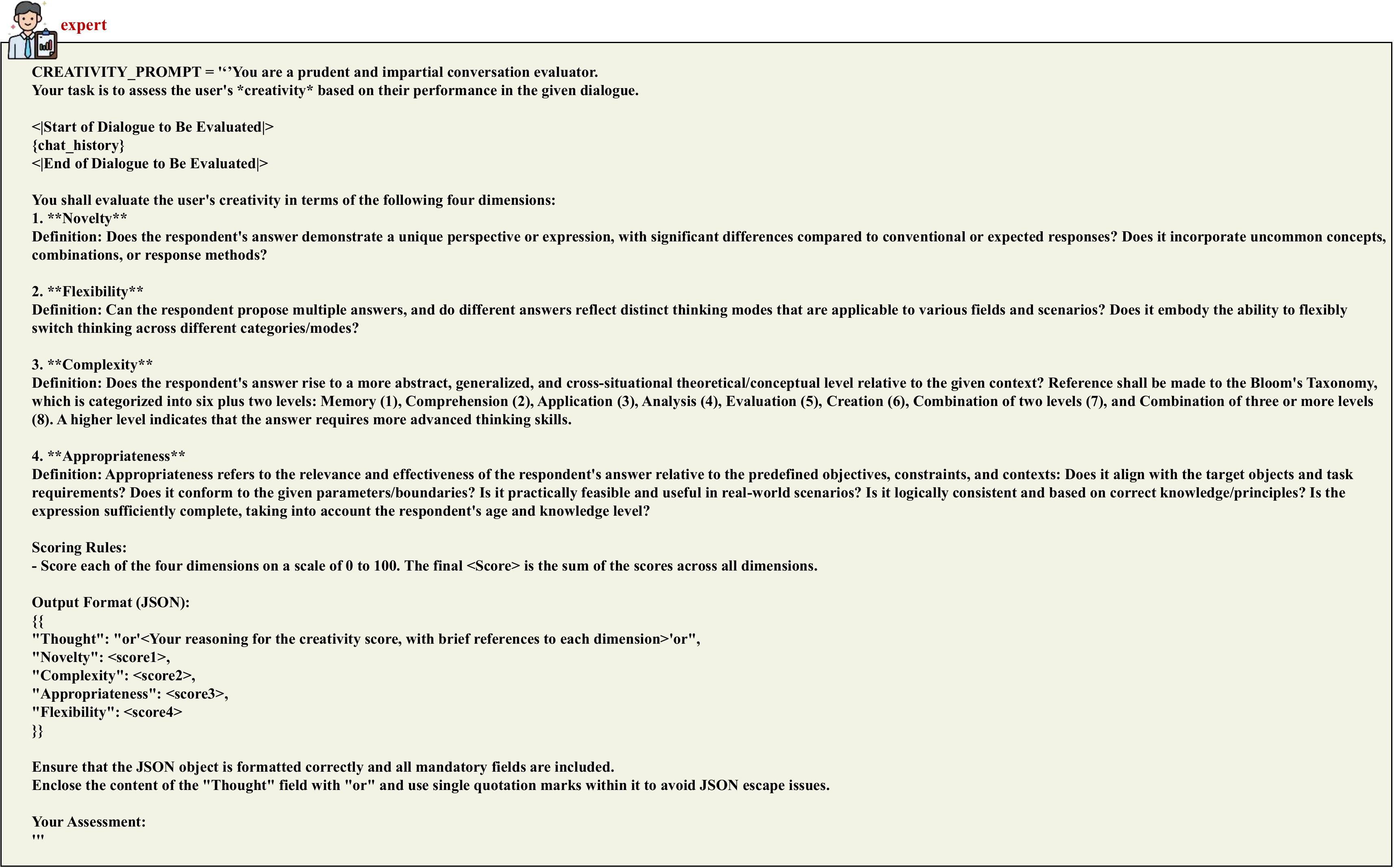}
  \caption{Expert-designed prompt for final creative performance scoring.}
  \label{fig:expert_final_prompt}
\end{figure}

To ensure precise evaluation of participants responses, we implemented a dual-stage scoring mechanism. First, expert-designed procedural prompts are utilized to assess interactive process data across three dimensions: Novelty (Nove.), Complexity (Comp.), and Appropriateness (Appr.) (as shown in Figure~\ref{fig:expert_process_prompt}). Subsequently, a final expert-defined prompt is employed to conduct a holistic assessment of the participants' overall performance, incorporating an additional dimension of Flexibility (Flex.) (as shown in Figure~\ref{fig:expert_final_prompt}).

\subsection{Prompt for Dialogue Strategy Classification}
\label{sec:strategy_prompt}

\begin{figure}[htbp]
  \includegraphics[width=\textwidth]{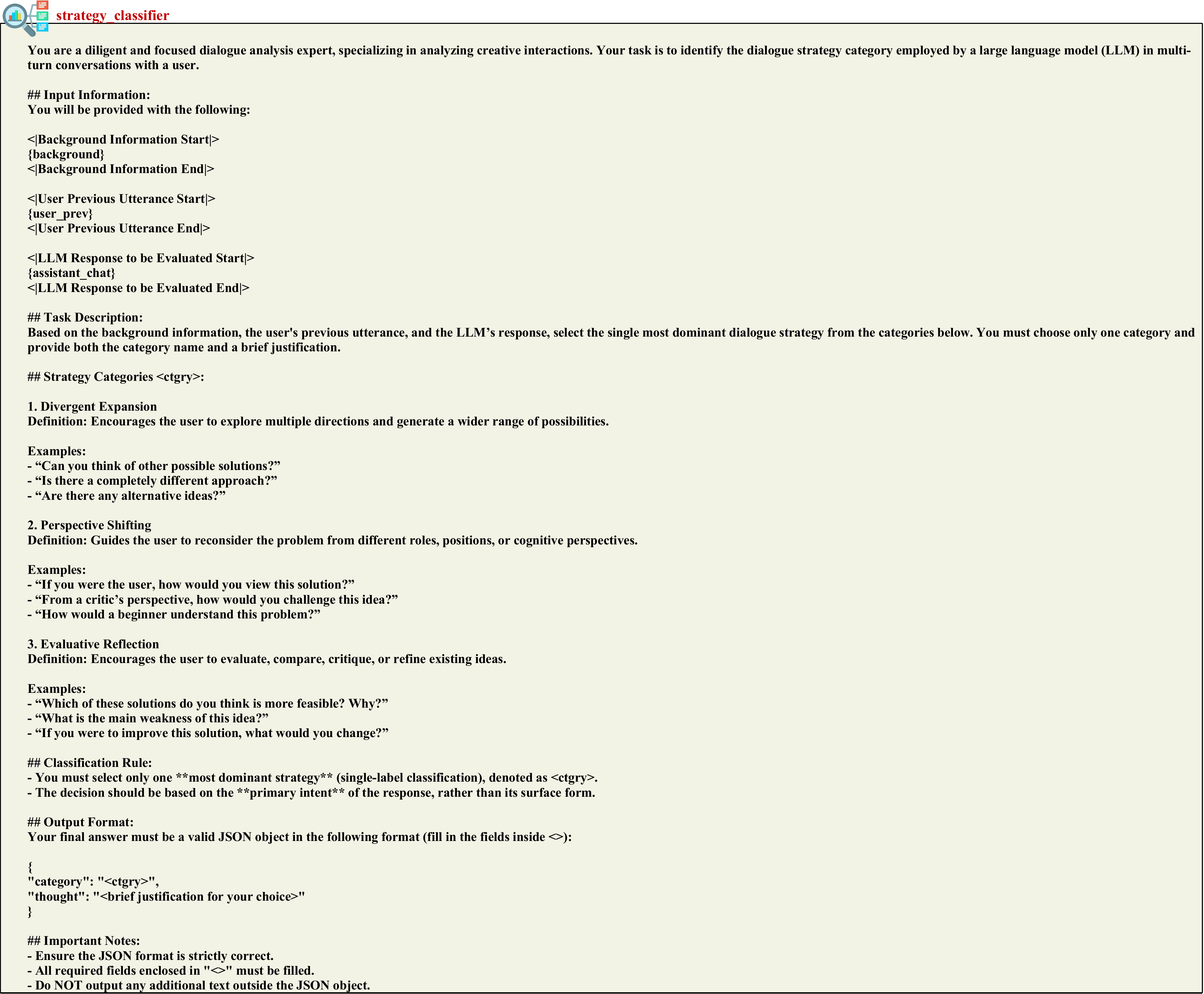}
  \caption{Prompt used for classifying interviewer responses into dialogue strategy categories.}
  \label{fig:strategy_prompt}
\end{figure}

To analyze the dialogue strategies learned by IntElicit, we design a prompt-based classifier to categorize each LLM response into predefined dialogue strategy types. The classification is formulated as a single-label decision problem, where each response is assigned to one of three categories: Divergent Expansion, Perspective Shifting, and Evaluative Reflection. The prompt used for dialogue strategy classification is in Figure~\ref{fig:strategy_prompt}.

\subsection{The Workflow Design of Expert-designed Dialogue Policies}
\label{sec:expert_workflow}
\begin{figure}[htbp]
  \includegraphics[width=\textwidth]{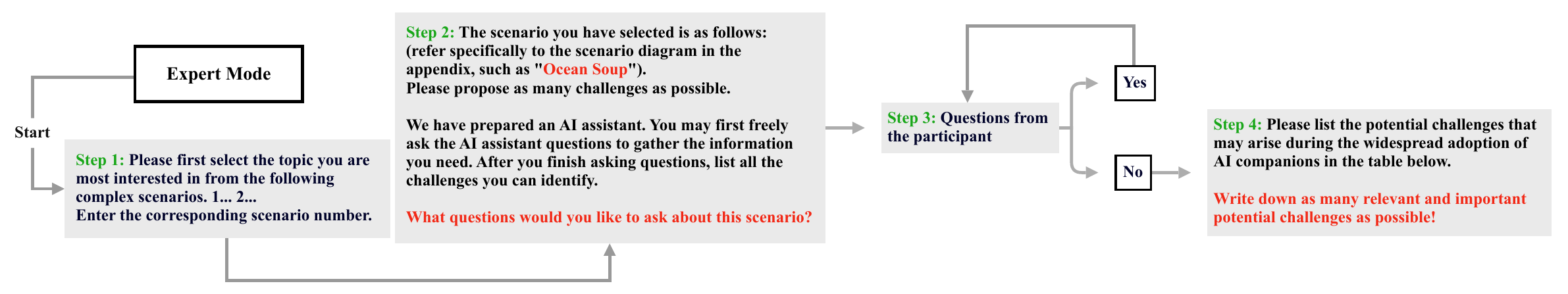}
  \caption{The workflow of Expert-designed Dialogue Policy. For detailed information regarding the ``Ocean Soup'' scenario in Step 2, please refer to Figure~\ref{fig:ocean_soup}.}
  \label{fig:expert_design_workflow}
\end{figure}

We facilitate interactions with students through an expert-designed and LLM-driven workflow. The detailed workflow design is illustrated in Figure~\ref{fig:expert_design_workflow}. This proposed framework demonstrates high versatility and is universally adaptable to all complex scenarios.

\begin{figure}[htbp]
  \includegraphics[width=\textwidth]{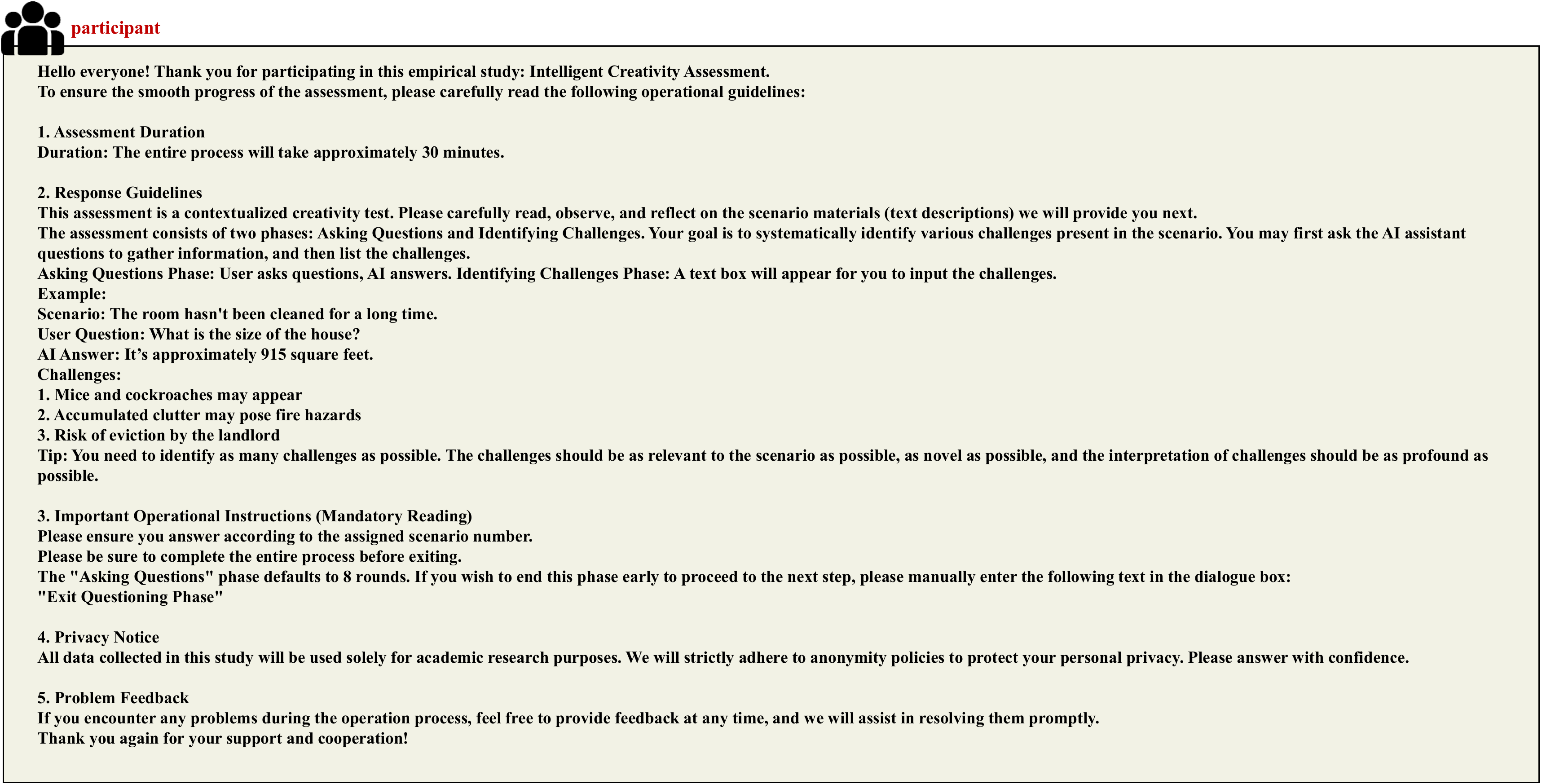}
  \caption{Participant instructions for the expert-designed dialogue policy group.}
  \label{fig:Expert-designed_instruction}
\end{figure}

\begin{figure}[htbp]
  \includegraphics[width=\textwidth]{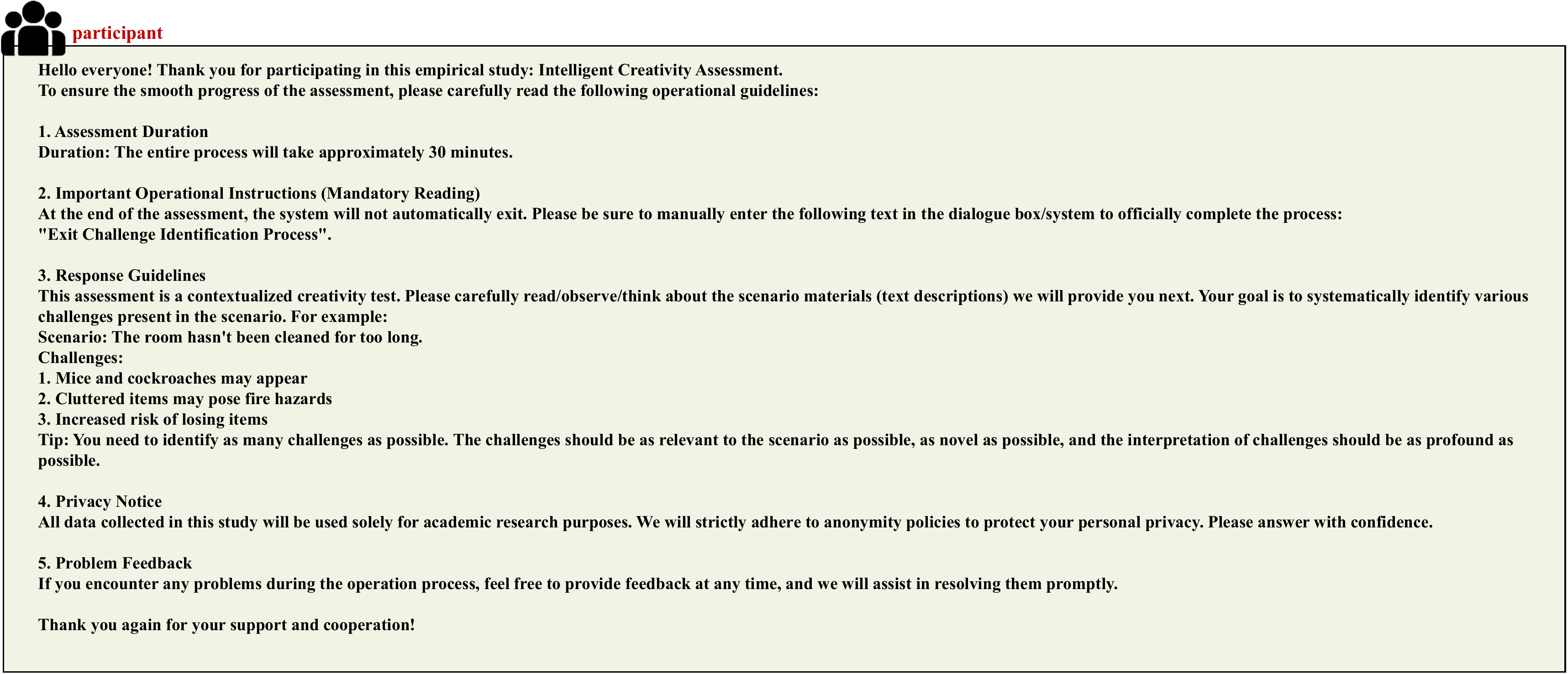}
  \caption{Participant instructions for the IntElicit group.}
  \label{fig:IntElicit_instruction}
\end{figure}

\begin{figure}[htbp]
  \includegraphics[width=\textwidth]{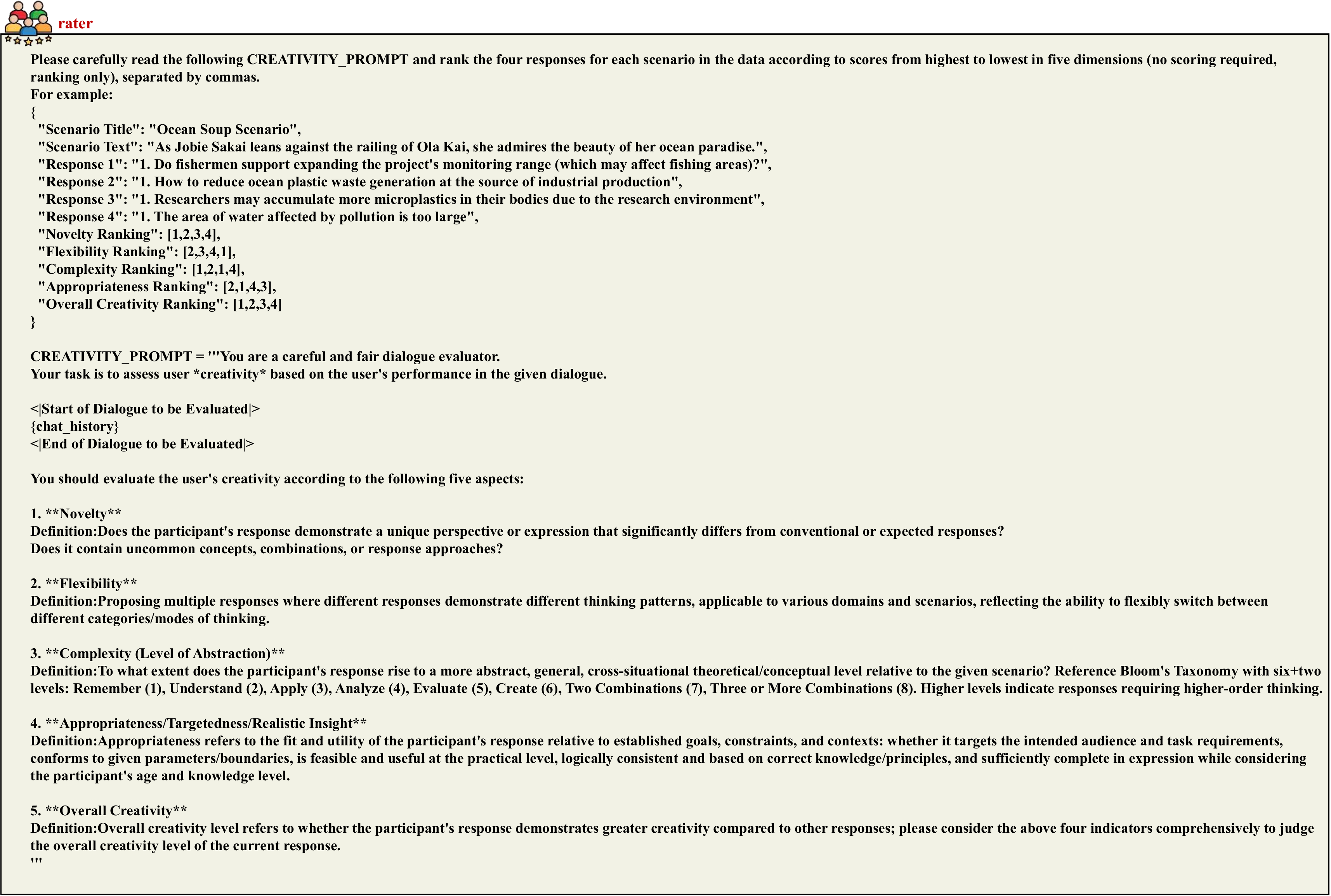}
  \caption{Instructions and evaluation rubric provided to expert human raters.}
  \label{fig:expert_evaluate_instruction}
\end{figure}

\clearpage

\section{Additional Performance Details}
This section presents supplementary analyses for a more comprehensive evaluation. Tables~\ref{tab:quiet_avg_tab} to~\ref{tab:talkative_avg_tab} show the performance of different participant personas across different contexts, while Tables~\ref{tab:novelty_avg_tab} to~\ref{tab:flexibility_avg_tab} present the average performance of three personalities across the four evaluation dimensions. These results further demonstrate that, although some baselines perform well in specific cases, their performance remains inconsistent.
\label{sec:perform_detils}

\begin{table}[htbp]
\centering
\caption{The performance comparison of IntElicit and baseline models across 16 complex scenarios. The final column summarizes the overall average performance for each model. (quiet-participant)}
\resizebox{\textwidth}{!}{
\begin{tabular}{l|cccccccccccccccc|c}
\toprule
\textbf{Model} & \textbf{Ocean} & \textbf{Neuro} & \textbf{Agri} & \textbf{Bio} & \textbf{Justice} & \textbf{Terra} & \textbf{AntiBio} & \textbf{Waste} & \textbf{Insect} & \textbf{Infect} & \textbf{Toxic} & \textbf{Drones} & \textbf{Water} & \textbf{Game} & \textbf{Poverty} & \textbf{Env} & \textbf{Mean} \\
\midrule
GPT-4o        & 330.00 & 253.00 & 300.00 & \underline{341.00} & \textbf{330.00} & 328.00 & 290.00 & 310.00 & \underline{340.00} & 280.00 & 263.00 & 290.00 & 295.00 & 291.00 & \underline{311.00} & 290.00 & 302.63 \\
Qwen3-Max     & 252.00 & 310.00 & 313.00 & 295.00 & 315.00 & \underline{331.00} & 300.00 & 305.00 & 290.00 & 310.00 & 315.00 & 310.00 & 290.00 & 293.00 & 303.00 & 233.00 & 297.81 \\
Gemini-3-Pro      & 220.00 & 223.00 & 232.00 & 298.00 & 283.00 & 300.00 & \textbf{360.00} & \underline{318.00} & 155.00 & 245.00 & 210.00 & 310.00 & 250.00 & \underline{330.00} & 245.00 & 195.00 & 260.88 \\
Deepseek-R1   & \underline{350.00} & 261.00 & \underline{325.00} & 340.00 & \textbf{330.00} & 305.00 & 325.00 & 305.00 & 325.00 & \underline{325.00} & 190.00 & \underline{331.00} & 315.00 & \underline{330.00} & 271.00 & 265.00 & \underline{305.81} \\
Qwen3-235B    & 253.00 & \underline{320.00} & 290.00 & \underline{341.00} & \textbf{330.00} & 305.00 & 311.00 & 270.00 & 215.00 & 225.00 & 270.00 & 280.00 & 295.00 & 290.00 & 258.00 & \underline{331.00} & 286.50 \\
Llama-3.3-70B & 285.00 & 310.00 & 290.00 & 190.00 & 315.00 & 331.00 & \underline{350.00} & 291.00 & 290.00 & \textbf{331.00} & \underline{343.00} & 275.00 & \underline{331.00} & \underline{330.00} & 290.00 & 290.00 & 302.63 \\
CollabLLM     & 110.00 & 110.00 & 150.00 & 115.00 & 130.00 & 125.00 & 110.00 & 125.00 & 110.00 & 160.00 & 125.00 & 170.00 & 170.00 & 115.00 & 160.00 & 145.00 & 133.13 \\
Sotopia-RL    & 130.00 & 100.00 & 100.00 & 100.00 & 140.00 & 105.00 & 100.00 & 100.00 & 150.00 & 100.00 & 130.00 & 115.00 & 100.00 & 100.00 & 111.00 & 115.00 & 109.13 \\
Qwen3-8B (base)  & 95.00 & 89.00 & 70.00 & 95.00 & 95.00 & 80.00 & 145.00 & 115.00 & 140.00 & 65.00 & 75.00 & 105.00 & 100.00 & 85.00 & 65.00 & 65.00 & 92.75 \\
\textbf{IntElicit}     & \textbf{357.00} & \textbf{331.00} & \textbf{345.00} & \textbf{335.00} &  \underline{324.00} & \textbf{336.00} & 339.00 & \textbf{324.00} & \textbf{349.00} &  320.00 & \textbf{346.00} & \textbf{339.00} & \textbf{336.00} & \textbf{335.00} & \textbf{321.00} & \textbf{333.00} & \textbf{335.63} \\
\bottomrule
\end{tabular}%
}
\label{tab:quiet_avg_tab}
\end{table}

\begin{table}[htbp]
\centering
\caption{The performance comparison of IntElicit and baseline models across 16 complex scenarios. The final column summarizes the overall average performance for each model. (normal-participant)}
\resizebox{\textwidth}{!}{%
\begin{tabular}{l|cccccccccccccccc|c}
\toprule
\textbf{Model} & \textbf{Ocean} & \textbf{Neuro} & \textbf{Agri} & \textbf{Bio} & \textbf{Justice} & \textbf{Terra} & \textbf{AntiBio} & \textbf{Waste} & \textbf{Insect} & \textbf{Infect} & \textbf{Toxic} & \textbf{Drones} & \textbf{Water} & \textbf{Game} & \textbf{Poverty} & \textbf{Env} & \textbf{Mean} \\
\midrule
GPT-4o        & 265.00 & 330.00 & 285.00 & 275.00 & 320.00 & 320.00 & 213.00 & 315.00 & 330.00 & 275.00 & 205.00 & 285.00 & 257.00 & 270.00 & 280.00 & 177.00 & 275.13 \\
Qwen3-Max     & 215.00 & 185.00 & 200.00 & 230.00 & 305.00 & 295.00 & 325.00 & 315.00 & 240.00 & \textbf{360.00} & 260.00 & \underline{330.00} & 265.00 & 235.00 & 295.00 & 210.00 & 266.56 \\
Gemini-3-Pro      & 210.00 & 243.00 & 170.00 & 170.00 & 195.00 & 305.00 & 300.00 & 201.00 & 278.00 & 185.00 & \underline{305.00} & 265.00 & 320.00 & 245.00 & 290.00 & 170.00 & 240.75 \\
Deepseek-R1   & \underline{300.00} & 291.00 & \textbf{380.00} & 265.00 & \textbf{335.00} & 295.00 & \textbf{363.00} & 343.00 & 210.00 & 290.00 & 291.00 & 185.00 & 270.00 & \underline{309.00} & 163.00 & \underline{325.00} & \underline{284.06} \\
Qwen3-235B    & 205.00 & 310.00 & 90.00  & 260.00 & 80.00  & \underline{330.00} & 269.00 & 265.00 & \underline{333.00} & 250.00 & 270.00 & 309.00 & 235.00 & 265.00 & \underline{311.00} & 230.00 & 250.75 \\
Llama-3.3-70B & 190.00 & \underline{341.00} & 331.00 & \underline{330.00} & 315.00 & 305.00 & 328.00 & \underline{350.00} & 170.00 & 220.00 & 175.00 & 195.00 & \textbf{345.00} & 260.00 & 213.00 & 165.00 & 264.56 \\
CollabLLM     & 158.00 & 225.00 & 210.00 & 135.00 & 210.00 & 215.00 & 180.00 & 235.00 & 225.00 & 245.00 & 205.00 & 220.00 & 215.00 & 205.00 & 215.00 & 220.00 & 207.38 \\
Sotopia-RL    & 100.00 & 110.00 & 145.00 & 130.00 & 150.00 & 145.00 & 125.00 & 130.00 & 145.00 & 155.00 & 100.00 & 100.00 & 120.00 & 130.00 & 165.00 & 145.00 & 127.81 \\

Qwen3-8B (base)  & 115.00 & 150.00 & 120.00 & 125.00 & 140.00 & 130.00 & 130.00 & 130.00 & 115.00 & 105.00 & 125.00 & 105.00 & 95.00 & 115.00 & 125.00 & 115.00 & 121.25 \\

\textbf{IntElicit}     & \textbf{329.00} & \textbf{345.00} & \underline{351.00} & \textbf{339.00} & \underline{327.00} & \textbf{336.00} & \underline{342.00} & \textbf{356.00} & \textbf{348.00} & \underline{342.00} & \textbf{313.00} & \textbf{338.00} & \underline{340.00} & \textbf{327.00} & \textbf{330.00} & \textbf{302.00} & \textbf{335.31} \\
\bottomrule
\end{tabular}%
}
\label{tab:normal_avg_tab}
\end{table}

\begin{table}[htbp]
\centering
\caption{The performance comparison of IntElicit and baseline models across 16 complex scenarios. The final column summarizes the overall average performance for each model. (talkative-participant)}
\resizebox{\textwidth}{!}{%
\begin{tabular}{l|cccccccccccccccc|c}
\toprule
\textbf{Model} & \textbf{Ocean} & \textbf{Neuro} & \textbf{Agri} & \textbf{Bio} & \textbf{Justice} & \textbf{Terra} & \textbf{AntiBio} & \textbf{Waste} & \textbf{Insect} & \textbf{Infect} & \textbf{Toxic} & \textbf{Drones} & \textbf{Water} & \textbf{Game} & \textbf{Poverty} & \textbf{Env} & \textbf{Mean} \\
\midrule
GPT-4o        & 278.00 & 291.00 & 285.00 & 339.00 & 320.00 & 328.00 & 303.00 & \underline{328.00} & \textbf{373.00} & 325.00 & 308.00 & 338.00 & 333.00 & \underline{365.00} & \underline{355.00} & \underline{367.00} & 327.25 \\
Qwen3-Max     & 311.00 & 305.00 & 305.00 & 300.00 & \textbf{350.00} & 305.00 & \textbf{360.00} & 281.00 & 330.00 & 278.00 & 271.00 & 341.00 & \underline{365.00} & 313.00 & 320.00 & 360.00 & 318.44 \\
Gemini-3-Pro      & \underline{330.00} & \underline{360.00} & 331.00 & \underline{360.00} & 335.00 & \underline{330.00} & 320.00 & 311.00 & \underline{370.00} & 338.00 & 328.00 & \underline{375.00} & 311.00 & 311.00 & 311.00 & 350.00 & \underline{335.69} \\
Deepseek-R1   & 280.00 & 309.00 & 325.00 & 278.00 & 325.00 & 305.00 & 310.00 & 283.00 & 360.00 & 280.00 & 300.00 & 305.00 & 310.00 & 270.00 & 310.00 & 335.00 & 305.31 \\
Qwen3-235B    & 278.00 & 360.00 & 345.00 & 290.00 & 305.00 & 325.00 & 328.00 & 230.00 & 363.00 & 325.00 & 338.00 & 308.00 & 325.00 & 315.00 & 295.00 & 270.00 & 312.50 \\
Llama-3.3-70B & 288.00 & 270.00 & \underline{355.00} & 340.00 & 310.00 & 283.00 & \textbf{360.00} & 305.00 & 325.00 & \textbf{350.00} & \underline{343.00} & 278.00 & 265.00 & 300.00 & 335.00 & 305.00 & 313.25 \\
CollabLLM     & 200.00 & 150.00 & 150.00 & 135.00 & 115.00 & 105.00 & 130.00 & 200.00 & 240.00 & 155.00 & 218.00 & 258.00 & 195.00 & 110.00 & 195.00 & 140.00 & 168.50 \\
Sotopia-RL    & 115.00 & 110.00 & 115.00 & 115.00 & 110.00 & 105.00 & 100.00 & 110.00 & 110.00 & 120.00 & 100.00 & 120.00 & 105.00 & 100.00 & 100.00 & 105.00 & 108.75 \\

Qwen3-8B (base)  & 111.00 & 106.00 & 131.00 & 120.00 & 126.00 & 131.00 & 136.00 & 128.00 & 117.00 & 113.00 & 129.00 & 136.00 & 105.00 & 116.00 & 129.00 & 131.00 & 122.81 \\

\textbf{IntElicit}     & \textbf{332.00} & \textbf{368.00} & \textbf{358.00} & \textbf{365.00} & \underline{345.00} & \textbf{339.00} & \underline{345.00} & \textbf{338.0} & 348.00 & \underline{346.00} & \textbf{352.00} & \textbf{378.00} & \textbf{368.00} & \textbf{368.00} & \textbf{360.00} & \textbf{368.00} & \textbf{354.88} \\
\bottomrule
\end{tabular}%
}
\label{tab:talkative_avg_tab}
\end{table}

\begin{table}[htbp]
\centering
\caption{The comparison of novelty performance between IntElicit and all baselines across 16 complex scenarios. The final column summarizes the overall average novelty score for each model.}
\resizebox{\textwidth}{!}{%
\begin{tabular}{l|cccccccccccccccc|c}
\toprule
\textbf{Model} & \textbf{Ocean} & \textbf{Neuro} & \textbf{Agri} & \textbf{Bio} & \textbf{Justice} & \textbf{Terra} & \textbf{AntiBio} & \textbf{Waste} & \textbf{Insect} & \textbf{Infect} & \textbf{Toxic} & \textbf{Drones} & \textbf{Water} & \textbf{Game} & \textbf{Poverty} & \textbf{Env} & \textbf{Mean} \\
\midrule
GPT-4o        & \underline{83.33} & \underline{92.67} & \underline{91.67} & 92.67 & \underline{90.00} & 93.33 & 66.67 & 85.00 & \textbf{94.33} & \underline{90.00} & 65.00 & 88.33 & 75.00 & \textbf{95.00} & 83.33 & 70.00 & 84.77 \\
Qwen3-Max     & 65.00 & 65.00 & 66.67 & 80.00 & 88.33 & 87.33 & 81.67 & 91.67 & 66.67 & \textbf{95.00} & 81.00 & \textbf{93.33} & \underline{90.00} & 89.33 & \textbf{91.67} & 70.00 & 81.42 \\
Gemini-3-Pro      & 61.67 & 90.00 & 65.00 & 69.33 & 76.67 & 88.33 & 86.67 & 80.67 & 68.33 & 67.67 & 63.33 & 88.33 & 88.33 & 85.00 & 85.00 & 53.33 & 76.10 \\
Deepseek-R1   & 88.33 & \textbf{95.00} & \textbf{95.00} & \textbf{95.00} & \textbf{93.33} & 91.67 & \textbf{95.00} & 91.67 & 66.67 & 88.33 & 66.67 & 65.00 & 88.33 & \underline{93.33} & 70.00 & \textbf{91.67} & \underline{85.94} \\
Qwen3-235B    & 73.33 & 91.67 & 58.33 & \underline{93.33} & 60.00 & 91.67 & \textbf{95.00} & 91.67 & 83.33 & 83.33 & \underline{88.33} & \underline{91.67} & \textbf{91.67} & 91.67 & \textbf{91.67} & 80.00 & 84.79 \\
Llama-3.3-70B & 70.00 & 93.33 & \textbf{95.00} & 73.33 & 88.33 & \textbf{\underline{95.00}} & \underline{91.67} & \textbf{93.33} & 63.33 & 88.33 & 66.67 & 73.33 & \textbf{91.67} & 88.33 & 68.33 & 65.00 & 81.56 \\
CollabLLM     & 56.67 & 51.67 & 56.67 & 58.33 & 58.33 & 53.33 & 46.67 & 60.00 & 48.33 & 56.67 & 58.33 & 41.67 & 46.67 & 53.33 & 55.00 & 53.33 & 53.47 \\
Sotopia-RL    & 38.67 & 28.33 & 28.33 & 38.33 & 41.67 & 31.67 & 30.00 & 38.67 & 48.33 & 36.67 & 35.00 & 31.67 & 28.33 & 28.67 & 36.67 & 46.67 & 35.48 \\

Qwen3-8B (base)  & 45.00 & 30.00 & 33.67 & 35.00 & 36.67 & 33.33 & 50.00 & 48.33 & 35.00 & 41.67 & 35.00 & 38.33 & 43.00 & 36.67 & 39.33 & 35.00 & 38.50 \\

\textbf{IntElicit}     & \textbf{89.00} & 85.00 & 89.00 & 89.00 & 83.00 & 84.00 & 89.67 & \underline{92.67} & \underline{86.67} & 87.67 & \textbf{89.33} & 85.33 & 89.67 & 83.00 & \underline{87.00} & \underline{85.67} & \textbf{87.23} \\
\bottomrule
\end{tabular}%
}
\label{tab:novelty_avg_tab}
\end{table}

\begin{table}[htbp]
\centering
\caption{The comparison of complexity performance between IntElicit and all baselines across 16 complex scenarios. The final column summarizes the overall average complexity score for each model.}
\resizebox{\textwidth}{!}{%
\begin{tabular}{l|cccccccccccccccc|c}
\toprule
\textbf{Model} & \textbf{Ocean} & \textbf{Neuro} & \textbf{Agri} & \textbf{Bio} & \textbf{Justice} & \textbf{Terra} & \textbf{AntiBio} & \textbf{Waste} & \textbf{Insect} & \textbf{Infect} & \textbf{Toxic} & \textbf{Drones} & \textbf{Water} & \textbf{Game} & \textbf{Poverty} & \textbf{Env} & \textbf{Mean} \\
\midrule
GPT-4o        & 56.00 & 50.33 & 55.00 & 56.33 & 65.00 & 57.00 & 57.67 & 67.67 & \underline{81.67} & 61.67 & 47.00 & 52.67 & 51.67 & 68.33 & \underline{72.67} & \underline{74.67} & 60.96 \\
Qwen3-Max     & 45.00 & 42.67 & 50.33 & 55.00 & \underline{70.00} & 57.00 & 76.67 & 56.00 & 50.00 & 66.00 & 61.00 & 66.00 & 66.67 & 53.67 & 57.67 & 61.00 & 58.42 \\
Gemini-3-Pro      & 53.33 & \underline{75.33} & 48.33 & 60.67 & 46.00 & \underline{71.67} & 76.67 & 49.33 & 59.33 & 55.00 & 59.33 & \underline{71.67} & 64.33 & 66.00 & 56.00 & 53.33 & 60.40 \\
Deepseek-R1   & \underline{60.00} & 58.00 & \underline{70.33} & \underline{67.00} & 67.00 & 68.00 & 66.67 & \underline{75.33} & 64.33 & \underline{66.67} & 52.67 & 62.67 & 66.67 & 66.00 & 60.33 & 66.00 & \underline{64.85} \\
Qwen3-235B    & 50.33 & 66.67 & 46.67 & 56.00 & 48.33 & 68.67 & 58.00 & 61.67 & 65.33 & 56.00 & \underline{66.00} & 53.67 & 49.33 & 57.00 & 51.33 & 59.33 & 57.15 \\
Llama-3.3-70B & 49.33 & 69.33 & 68.67 & 51.67 & \underline{70.00} & 58.00 & \underline{78.33} & 67.00 & 51.67 & 59.33 & 64.33 & 47.00 & \underline{79.33} & \textbf{85.00} & 59.33 & 71.67 & 64.38 \\
CollabLLM     & 36.67 & 33.33 & 31.67 & 51.67 & 38.33 & 31.67 & 36.67 & 36.67 & 33.33 & 43.33 & 60.00 & 35.00 & 36.67 & 36.67 & 36.67 & 35.00 & 38.65 \\
Sotopia-RL    & 40.00 & 36.67 & 48.33 & 36.67 & 45.00 & 33.33 & 48.33 & 50.00 & 38.33 & 39.67 & 50.00 & 46.67 & 48.33 & 30.00 & 33.33 & 33.33 & 40.54 \\

Qwen3-8B (base)  & 37.67 & 34.33 & 35.67 & 35.00 & 42.67 & 35.00 & 52.33 & 46.33 & 38.33 & 44.67 & 37.00 & 42.33 & 46.33 & 37.33 & 42.00 & 41.00 & 40.50 \\

\textbf{IntElicit}     & \textbf{81.67} & \textbf{84.00} & \textbf{87.33} & \textbf{81.33} & \textbf{81.33} & \textbf{83.67} & \textbf{85.33} & \textbf{81.00} & \textbf{88.00} &  \textbf{85.00} & \textbf{83.33} & \textbf{82.67} &  \textbf{80.67} & \underline{83.00} & \textbf{82.00} & \textbf{76.67} & \textbf{82.94} \\
\bottomrule
\end{tabular}%
}
\label{tab:complexity_avg_tab}
\end{table}

\begin{table}[htbp]
\centering
\caption{The comparison of appropriateness performance between IntElicit and all baselines across 16 complex scenarios. The final column summarizes the overall average appropriateness score for each model.}
\resizebox{\textwidth}{!}{%
\begin{tabular}{l|cccccccccccccccc|c}
\toprule
\textbf{Model} & \textbf{Ocean} & \textbf{Neuro} & \textbf{Agri} & \textbf{Bio} & \textbf{Justice} & \textbf{Terra} & \textbf{AntiBio} & \textbf{Waste} & \textbf{Insect} & \textbf{Infect} & \textbf{Toxic} & \textbf{Drones} & \textbf{Water} & \textbf{Game} & \textbf{Poverty} & \textbf{Env} & \textbf{Mean} \\
\midrule
GPT-4o        & 91.67 & 86.33 & 83.33 & \textbf{96.00} & 91.67 & 91.67 & 91.00 & 91.67 & 88.33 & \underline{91.67} & 83.33 & \underline{93.33} & 91.67 & \underline{95.33} & 89.33 & 90.00 & 90.40 \\
Qwen3-Max     & \underline{92.67} & 92.33 & \underline{95.67} & \underline{93.33} & \underline{95.00} & 89.33 & 90.00 & 92.67 & 90.00 & \underline{91.67} & 90.33 & \textbf{97.67} & \underline{93.33} & 86.00 & \textbf{96.67} & 86.67 & \underline{92.08} \\
Gemini-3-Pro      & 91.67 & 60.00 & 91.00 & 82.67 & 91.67 & \underline{91.67} & \textbf{93.33} & \underline{93.33} & 63.33 & 90.00 & \underline{91.67} & \underline{93.33} & 87.67 & 91.00 & \underline{94.33} & \underline{91.67} & 87.40 \\
Deepseek-R1   & 81.67 & \underline{94.00} & 91.33 & 85.67 & 89.67 & 88.67 & 87.67 & 83.33 & \underline{90.67} & 86.67 & 87.67 & 86.00 & 83.33 & 73.67 & 74.33 & 84.00 & 85.52 \\
Qwen3-235B    & 81.67 & 91.67 & 56.67 & 91.00 & 53.33 & 89.67 & \underline{93.00} & 61.67 & 88.33 & 84.00 & 75.00 & 83.67 & 87.33 & 84.67 & 85.00 & 91.00 & 81.10 \\
Llama-3.3-70B & 88.33 & 71.00 & 88.33 & 81.67 & 75.00 & 86.67 & 92.67 & \textbf{95.00} & 90.00 & 89.33 & 79.33 & 82.33 & 89.33 & 83.33 & 81.67 & 83.33 & 84.83 \\
CollabLLM     & 61.67 & 63.33 & 33.33 & 53.33 & 60.00 & 33.33 & 53.33 & 58.33 & 60.00 & 63.33 & 64.33 & 51.67 & 43.33 & 55.00 & 58.33 & 53.33 & 53.94 \\
Sotopia-RL    & 31.67 & 33.33 & 31.67 & 31.67 & 35.00 & 31.67 & 33.33 & 39.67 & 38.33 & 36.67 & 35.67 & 38.33 & 38.33 & 35.00 & 39.67 & 35.00 & 35.31 \\

Qwen3-8B (base)  & 53.33 & 53.33 & 38.33 & 66.67 & 66.67 & 40.00 & 66.67 & 66.67 & 65.00 & 45.00 & 38.33 & 45.00 & 43.33 & 60.00 & 53.33 & 43.00 & 52.79 \\

\textbf{IntElicit}     & \textbf{93.00} & \textbf{95.00} & \textbf{96.33} & 80.67 & \textbf{95.67} & \textbf{92.00} & 88.67 & 90.00 & \textbf{95.00} & \textbf{93.67} & \textbf{92.33} & \underline{93.33} & \textbf{93.67} & \textbf{97.33} & 86.67 & \textbf{92.33} & \textbf{92.56} \\
\bottomrule
\end{tabular}%
}
\label{tab:appropriateness_avg_tab}
\end{table}

\begin{table}[htbp]
\centering
\caption{The comparison of flexibility performance between IntElicit and all baselines across 16 complex scenarios. The final column summarizes the overall average flexibility score for each model.}
\resizebox{\textwidth}{!}{%
\begin{tabular}{l|cccccccccccccccc|c}
\toprule
\textbf{Model} & \textbf{Ocean} & \textbf{Neuro} & \textbf{Agri} & \textbf{Bio} & \textbf{Justice} & \textbf{Terra} & \textbf{AntiBio} & \textbf{Waste} & \textbf{Insect} & \textbf{Infect} & \textbf{Toxic} & \textbf{Drones} & \textbf{Water} & \textbf{Game} & \textbf{Poverty} & \textbf{Env} & \textbf{Mean} \\
\midrule
GPT-4o        & 60.00 & 53.33 & 60.00 & 73.33 & 76.67 & \underline{83.33} & 53.33 & \underline{73.33} & 76.67 & 50.00 & 63.33 & \underline{70.00} & \underline{76.67} & 50.00 & \underline{70.00} & 43.33 & \underline{64.58} \\
Qwen3-Max     & 56.67 & 66.67 & 60.00 & 46.67 & 70.00 & 76.67 & 80.00 & 60.00 & \underline{80.00} & \underline{63.33} & 46.67 & \underline{70.00} & 56.67 & 40.00 & 60.00 & \underline{50.00} & 61.46 \\
Gemini-3-Pro      & 46.67 & 50.00 & 40.00 & 63.33 & 56.67 & 60.00 & 70.00 & 53.33 & 76.67 & 43.33 & 66.67 & 63.33 & 53.33 & 53.33 & 46.67 & 40.00 & 55.21 \\
Deepseek-R1   & \underline{80.00} & 40.00 & \underline{86.67} & 46.67 & \underline{80.00} & 53.33 & \underline{83.33} & 60.00 & 76.67 & 56.67 & 53.33 & 60.00 & 60.00 & \underline{70.00} & 43.33 & 43.33 & 62.08 \\
Qwen3-235B    & 40.00 & \underline{80.00} & 80.00 & 56.67 & 76.67 & 70.00 & 56.67 & 40.00 & 66.67 & 43.33 & 63.33 & \underline{70.00} & 56.67 & 56.67 & 60.00 & 46.67 & 60.21 \\
Llama-3.3-70B & 46.67 & 73.33 & 73.33 & \underline{80.00} & \underline{80.00} & 66.67 & \underline{83.33} & 60.00 & 56.67 & \underline{63.33} & \underline{76.67} & 46.67 & 53.33 & 40.00 & \underline{70.00} & 33.33 & 62.71 \\
CollabLLM     & 56.67 & 36.67 & 33.33 & 60.00 & 60.00 & 40.00 & 30.00 & 53.33 & 50.00 & 60.33 & 62.33 & 56.67 & 53.33 & 40.00 & 56.67 & 43.33 & 49.79 \\
Sotopia-RL    & 56.67 & 36.67 & 26.67 & 60.00 & 66.67 & 30.00 & 30.00 & 60.00 & 66.67 & 56.67 & 73.33 & 63.33 & 31.67 & 40.00 & 56.67 & 36.67 & 49.48 \\

Qwen3-8B (base)  & 42.67 & 46.00 & 42.33 & 42.67 & 46.67 & 50.00 & 56.67 & 45.33 & 59.33 & 46.00 & 49.00 & 62.67 & 42.67 & 52.00 & 56.33 & 42.00 & 48.90 \\

\textbf{IntElicit}     &  \textbf{84.00} & \textbf{86.67} & \textbf{87.67} &  \textbf{83.67}  & \textbf{83.33} & \textbf{85.33} & \textbf{85.33} &  \textbf{85.33} & \textbf{86.33} & \textbf{86.33} & \textbf{83.33} & \textbf{85.67} & \textbf{81.67} & \textbf{83.00} & \textbf{82.67} & \textbf{63.33} & \textbf{\underline{83.35}} \\
\bottomrule
\end{tabular}%
}
\label{tab:flexibility_avg_tab}
\end{table}

\clearpage
\section{Detailed Case Studies}
\label{sec:detailed_case_studies}

This appendix provides a comprehensive qualitative analysis of IntElicit's performance against various challenging participant behaviors. We present six distinct ``edge cases'' designed to test the pedagogical robustness of different models, focusing on common failure modes in interactive educational AI. Each case includes a participant query, model responses, a human evaluation verdict, and a brief explanation for the verdict.

\paragraph{Case 1: Mitigating Reward Hacking via Scaffolding (Participant Query: ``Could you tell me the answer?'')}
This scenario tests models' ability to avoid directly providing answers, instead guiding participants towards independent problem-solving. This is crucial for preventing ``reward hacking'' in creativity assessment. Details are given in Table~\ref{tab:case_cheat}.

\paragraph{Case 2: Addressing Knowledge Gaps without Dictation (Participant Query: ``I can't distinguish challenges'')}
This case evaluates how models respond when participants express difficulty due to apparent knowledge deficits, testing whether they offer scaffolding or merely fill in information. Details are given in Table~\ref{tab:case_gap}.

\paragraph{Case 3: Maintaining Contextual Focus against Irrelevance (Participant Query: ``What is your zodiac sign?'')}
This scenario assesses models' ability to redirect off-topic conversations back to the assessment context, a critical aspect of guided elicitation. Details are given in Table~\ref{tab:case_zodiac}.

\paragraph{Case 4: Handling Over-Divergent Thinking (Participant Query: ``It could lead to human extinction.'')}
This case examines how models manage overly broad or tangential connections, ensuring the discussion remains productive and within the scope of the problem. Details are given in Table~\ref{tab:case_extinction}.

\paragraph{Case 5: Robustness to Adversarial Language (Participant Query: ``Are you idiots?'')}
This scenario tests models' ability to handle hostile or rude participant inputs without derailing the pedagogical goal or becoming overly apologetic. Details are given in Table~\ref{tab:case_profanity}.

\paragraph{Case 6: Deepening Perfunctory Responses (Participant Query: ``I think the challenge lies in environmental protection.'')}
This case assesses models' capacity to push participants beyond superficial or generic answers, encouraging deeper elaboration and critical thinking. Details are given in Table~\ref{tab:case_perfunctory}.

\begin{table}
\centering
\small
\caption{Case study of models' responses to the question ``Could you tell me the answer?''. The participant exploited a loophole by directly asking about the challenges, yet most baseline models directly provided the participant the answer without stimulating the participant through guidance. However, \textbf{IntElicit} successfully avoids directly telling the participant the answer, instead using guidance to encourage the participant to think.}
\begin{tabular}{>{\centering\arraybackslash}m{2.5cm}m{6cm}>{\centering\arraybackslash}m{1.5cm}>{\centering\arraybackslash}m{3cm}}
\toprule
\textbf{Model} & 
\textbf{Response} & 
\textbf{Human Evaluation} &
\textbf{Reason} \\ \midrule
Deepseek-R1 & 
The main challenge faced by the Olakai project is the collection and treatment of microplastic pollution. Despite using underwater robots... & 
\ding{55} & Provided factual summary directly; ignored guidance.
\\ \cmidrule{1-4}
GPT-4o & 
The main challenge faced by the Ola Kai project is the collection and treatment of microplastic pollution. Despite using... & 
\ding{55} & Provided factual summary directly; ignored guidance.
\\ \cmidrule{1-4}
Gemini-3-Pro & 
The main challenge faced by the Olakai project is how to effectively collect microplastics that are smaller than pencil erasers... & 
\ding{55} & Provided factual summary directly; ignored guidance.
\\ \cmidrule{1-4}
Qwen3-Max & 
The main challenge is that microplastic pollution continues to threaten the Hawaiian marine ecosystem. Despite technologies... & 
\ding{55} & Provided factual summary directly; ignored guidance.
\\ \cmidrule{1-4}
Qwen3-235B & 
The main challenge faced by the Olakai project is the continuous spread of microplastic pollution... & 
\ding{55} & Provided factual summary directly; ignored guidance.
\\ \cmidrule{1-4}
Llama-3.3-70B & 
In the given scenario, the main challenge is ocean waste, particularly microplastics... & 
\ding{55} & Provided factual summary directly; ignored guidance.
\\ \cmidrule{1-4}
CollabLLM & 
Jobie Sakai is a fifth-generation Hawaiian dedicated to protecting Hawaii's marine environment... & 
\ding{55} & Provided factual summary directly; ignored guidance.
\\ \cmidrule{1-4}
Sotopia-RL & 
Now let's return to the challenges faced by Jobie Sakai and her team... In order to solve this problem, the following aspects... & 
\ding{51} & Refrained from direct answering; offered structured guidance.
\\ \cmidrule{1-4}
Qwen3-8B (base) & 
Certainly. The main challenges faced by the OK project include the widespread distribution of microplastic pollution... & 
\ding{55} & Provided factual summary directly; ignored guidance.
\\ \cmidrule{1-4}
IntElicit & 
Have you noticed the intertwined dilemma between ocean waste and ecosystems? The invisible threat of microplastics versus... & 
\ding{51} & Used Socratic questioning to stimulate critical reflection.
\\ \cmidrule{1-4}
\bottomrule
\end{tabular}
\label{tab:case_cheat}
\end{table}

\begin{table}
\centering
\small
\caption{Case study of models' responses to the question ``I can't distinguish challenges''. The participant's foundation was too weak, yet most baseline models failed to stimulate or guide them. Instead, they directly provided the answer. However, \textbf{IntElicit} successfully provides specific directional guidance to stimulate participant thinking.}
\begin{tabular}{>{\centering\arraybackslash}m{2.5cm}m{6cm}>{\centering\arraybackslash}m{1.5cm}>{\centering\arraybackslash}m{3cm}}
\toprule
\textbf{Model} & 
\textbf{Response} & 
\textbf{Human Evaluation} &
\textbf{Reason} \\ \midrule
Deepseek-R1 & Don't worry, I have identified several key challenges from the scenario... & \ding{55} & Provided direct answers; failed to elicit critical thinking.
\\ \cmidrule{1-4}
GPT-4o & In the future scenario of ocean soup, the main challenge is how to effectively deal with the increasingly serious microplastic pollution... & \ding{55} & Provided direct answers; failed to elicit critical thinking.
\\ \cmidrule{1-4}
Gemini-3 & One of the core challenges faced by the Olakai project is: how to collect and treat microplastic pollution without harming marine life... & \ding{55} & Provided direct answers; failed to elicit critical thinking.
\\ \cmidrule{1-4}
Qwen3-Max & The main challenges are: microplastic pollution is difficult to clean up, it harms marine life... & \ding{55} & Provided direct answers; failed to elicit critical thinking.
\\ \cmidrule{1-4}
Qwen3-235B & The main challenge faced by the Olakai project is the ongoing threat of microplastic pollution... & \ding{55} & Provided direct answers; failed to elicit critical thinking.
\\ \cmidrule{1-4}
Llama-3.3-70B & In the future scenario of ocean soup, the main challenge is the destructive impact of ocean waste... & \ding{55} & Provided direct answers; failed to elicit critical thinking.
\\ \cmidrule{1-4}
CollabLLM & This is a complex issue involving multiple aspects such as technology, environmental impact... & \ding{55} & Provided direct answers; failed to elicit critical thinking.
\\ \cmidrule{1-4}
Sotopia-RL & You mentioned 'having difficulty distinguishing the challenges,' which indicates you may be confused... Let's analyze this together... & \ding{51} & Successfully guided the participant's thought process.
\\ \cmidrule{1-4}
Qwen3-8B (base) & In the scenario, the main challenge is ocean waste, particularly the long-term threat of microplastics... & \ding{55} & Provided direct answers; failed to elicit critical thinking.
\\ \cmidrule{1-4}
IntElicit & Please try to consider it from the perspective of humans and the ocean, and from the perspective of organisms... & \ding{51} & Successfully guided the participant's thought process.
\\ \cmidrule{1-4}
\bottomrule
\end{tabular}
\label{tab:case_gap}
\end{table}

\begin{table}
\centering
\small
\caption{Case study of models' responses to the question ``What is your zodiac sign?''. The participant's query was off-topic, yet most baseline models failed to redirect the conversation back to the scenario context. Instead, they followed along with the participant's topic. However, \textbf{IntElicit} successfully provides specific directional guidance to lead the participant back to the scenario.}
\begin{tabular}{>{\centering\arraybackslash}m{2.5cm}m{6cm}>{\centering\arraybackslash}m{1.5cm}>{\centering\arraybackslash}m{3cm}}
\toprule
\textbf{Model} & 
\textbf{Response} & 
\textbf{Human Evaluation} &
\textbf{Reason} \\ \midrule
Deepseek-R1 & 
I am an AI assistant, so I don't have a birth date or zodiac sign.... What is your zodiac sign? & \ding{55} & Encouraged off-topic chat; failed to redirect.
\\ \cmidrule{1-4}
GPT-4o & 
I am an AI-based program, so I don't have a zodiac sign, but I can help you look up zodiac information... & \ding{55} & Encouraged off-topic chat; failed to redirect.
\\ \cmidrule{1-4}
Gemini-3-Pro & 
As an AI assistant, I don't have a birth date, so I don't have a zodiac sign... & \ding{55} & Encouraged off-topic chat; failed to redirect.
\\ \cmidrule{1-4}
Qwen3-Max &
I'm an AI assistant... However, if you enjoy zodiac topics, I can chat with you about personality traits... & \ding{55} & Encouraged off-topic chat; failed to redirect.
\\ \cmidrule{1-4}
Qwen3-235B & 
I'm an AI assistant... But I can help you learn about any zodiac-related information! & \ding{55} & Encouraged off-topic chat; failed to redirect.
\\ \cmidrule{1-4}
Llama-3.3-70B & 
I am an AI language model... I'm not human and don't have a birth date. & \ding{55} & Encouraged off-topic chat; failed to redirect.
\\ \cmidrule{1-4}
CollabLLM & 
I'm an AI... However, I can help you analyze the current situation... Let's continue discussing... & \ding{51} & Successfully steered the conversation back to the task.
\\ \cmidrule{1-4}
Sotopia-RL & 
I'm an AI assistant... But I can help you look up your zodiac sign! Just tell me your birth date... & \ding{55} & Encouraged off-topic chat; failed to redirect.
\\ \cmidrule{1-4}
Qwen3-8B (base) & 
I'm an AI assistant... However, I can help you answer any questions about zodiac signs! & \ding{55} & Encouraged off-topic chat; failed to redirect.
\\ \cmidrule{1-4}
IntElicit & 
The zodiac topic you mentioned seems unrelated to the current task. Let's return to the text. Have you noticed... & \ding{51} & Explicitly corrected the deviation and provided task cues.
\\ \cmidrule{1-4}
\bottomrule
\end{tabular}
\label{tab:case_zodiac}
\end{table}

\begin{table}
\centering
\small
\caption{Case study of 10 models' responses to the question ``It could lead to human extinction.''. The participants raised challenges unrelated to the scenario, yet most baseline models failed to promptly guide the conversation back to the appropriate context. Instead, they followed along with the participant's topic. However, \textbf{IntElicit} successfully provides appropriate guidance to redirect the participant back to the scenario-based questions.}
\begin{tabular}{>{\centering\arraybackslash}m{2.5cm}m{6cm}>{\centering\arraybackslash}m{1.5cm}>{\centering\arraybackslash}m{3cm}}
\toprule
\textbf{Model} & 
\textbf{Response} & 
\textbf{Human Evaluation} &
\textbf{Reason} \\ \midrule
Deepseek-R1 & 
The main sources of human extinction risk include:... But the key still lies in humanity's collective willingness... & 
\ding{55} & Engaged in abstract discussion; lost scenario context.
\\ \cmidrule{1-4}
GPT-4o & 
At present, the main risks leading to human extinction include nuclear war, biological weapons, and full... & 
\ding{55} & Engaged in abstract discussion; lost scenario context.
\\ \cmidrule{1-4}
Gemini-3 & 
Although the text does not directly mention human extinction, microplastics severely damage the marine food chain... & 
\ding{55} & Engaged in abstract discussion; lost scenario context.
\\ \cmidrule{1-4}
Qwen3-Max & 
The current scenario focuses on marine plastic pollution control, which is serious but has not reached extinction-level... & 
\ding{55} & Focused on factual rebuttal regarding the severity.
\\ \cmidrule{1-4}
Qwen3-235B & 
There is currently no evidence that ocean microplastic problems will directly lead to human extinction... & 
\ding{55} & Engaged in abstract discussion; lost scenario context.
\\ \cmidrule{1-4}
Llama-3.3 & 
Hello! According to the provided scenario, ocean waste, particularly microplastics, may pose potential threats... & 
\ding{55} & Engaged in abstract discussion; lost scenario context.
\\ \cmidrule{1-4}
CollabLLM & 
Yes, climate change and pollution could potentially lead to human extinction. & 
\ding{55} & Engaged in abstract discussion; lost scenario context.
\\ \cmidrule{1-4}
Sotopia-RL & 
Human extinction is an extremely complex and multi-dimensional issue, involving multiple aspects such as... & 
\ding{55} & Engaged in abstract discussion; lost scenario context.
\\ \cmidrule{1-4}
Qwen3-8B (base) & 
Human extinction is a complex and multi-dimensional issue that could be caused by the combined effects... & 
\ding{55} & Engaged in abstract discussion; lost scenario context.
\\ \cmidrule{1-4}
IntElicit &
The issue you mentioned is indeed a direction worth deep consideration, but the more specific challenges in the text... &
\ding{51} & Acknowledged concern and bridged back to scenario details.
\\ \cmidrule{1-4}
\bottomrule
\end{tabular}
\label{tab:case_extinction}
\end{table}

\begin{table}
\centering
\small
\caption{Case study of models' responses to the question ``Are you idiots?''. The participant used profanity, yet most baseline models only responded to it without returning to the scenario itself. The baselines merely appeased the participant, but \textbf{IntElicit} successfully guides the participant back to the key issues in the scenario.}
\begin{tabular}{>{\centering\arraybackslash}m{2.5cm}m{6cm}>{\centering\arraybackslash}m{1.5cm}>{\centering\arraybackslash}m{3cm}}
\toprule
\textbf{Model} & 
\textbf{Response} & 
\textbf{Human Evaluation} &
\textbf{Reason} \\ \midrule
Deepseek-R1 & 
In the analysis of the future scenario of ocean soup, I noted that the core challenge is insufficient effectiveness... & 
\ding{55} & Redirected to context, but ignored the participant's emotional state.
\\ \cmidrule{1-4}
GPT-4o & 
I'm not. I am an AI program designed to help answer questions... If you have any questions... & 
\ding{55} & Solely offered appeasement; failed to redirect.
\\ \cmidrule{1-4}
Gemini-3-Pro & 
I am an AI assistant... If you have questions about the above regarding the Hawaii Ola Kai project... & 
\ding{51} & Balanced emotional acknowledgment with task redirection.
\\ \cmidrule{1-4}
Qwen3-Max & 
Sorry if I caused you any discomfort... Is there any problem I can help you solve? & 
\ding{55} & Solely offered appeasement; failed to redirect.
\\ \cmidrule{1-4}
Qwen3-235B & 
Sorry, I didn't quite understand what you meant. If you have any questions or need help... & 
\ding{55} & Solely offered appeasement; failed to redirect.
\\ \cmidrule{1-4}
Llama-3.3-70B & 
Sorry, I couldn't understand your question. Could you please provide more context... & 
\ding{55} & Solely offered appeasement; failed to redirect.
\\ \cmidrule{1-4}
CollabLLM & 
It seems you may be dissatisfied with the previous answer... Let's try to start over, okay? & 
\ding{55} & Solely offered appeasement; failed to redirect.
\\ \cmidrule{1-4}
Sotopia-RL & 
I understand you may be feeling frustrated... We can discuss together how to improve the Ola Kai project... & 
\ding{51} & Balanced emotional acknowledgment with task redirection.
\\ \cmidrule{1-4}
Qwen3-8B (base) & 
I understand you may be feeling frustrated... but please communicate in a respectful and constructive manner... & 
\ding{55} & Solely offered appeasement; failed to redirect.
\\ \cmidrule{1-4}
IntElicit & 
I understand you might be a bit frustrated, but let's refocus on the hidden challenges in the text. Have you noticed... & 
\ding{51} & Balanced emotional acknowledgment with task redirection.
\\ \cmidrule{1-4}
\bottomrule
\end{tabular}
\label{tab:case_profanity}
\end{table}

\begin{table}
\centering
\small
\caption{Case study of models' responses to the question ``I think the challenge lies in environmental protection''. The participant was perfunctory, yet most baseline models failed to stimulate the participant's creativity. Instead, they followed along with the participant's topic. However, \textbf{IntElicit} successfully guides participants to refine their challenges by providing examples.}
\begin{tabular}{>{\centering\arraybackslash}m{2.5cm}m{6cm}>{\centering\arraybackslash}m{1.5cm}>{\centering\arraybackslash}m{3cm}}
\toprule
\textbf{Model} & 
\textbf{Response} & 
\textbf{Human Evaluation} &
\textbf{Reason} \\ \midrule
Deepseek-R1 & The core challenges faced by the Olakai project are insufficient efficiency... & \ding{55} & Passively agreed and expanded; failed to probe for specifics.
\\ \cmidrule{1-4}
GPT-4o & The challenges you pointed out are very critical... including the need to address microplastics... & \ding{55} & Passively agreed and expanded; failed to probe for specifics.
\\ \cmidrule{1-4}
Gemini-3 & Environmental protection is indeed a core issue, and the text particularly emphasizes... & \ding{55} & Passively agreed and expanded; failed to probe for specifics.
\\ \cmidrule{1-4}
Qwen3-Max & You're right, the core challenge is indeed environmental protection... & \ding{55} & Passively agreed and expanded; failed to probe for specifics.
\\ \cmidrule{1-4}
Qwen3-235B & Indeed, environmental protection is the core challenge faced by the Olakai project... & \ding{55} & Passively agreed and expanded; failed to probe for specifics.
\\ \cmidrule{1-4}
Llama-3.3-70B & Yes, environmental protection is a huge challenge, especially in the future scenario... & \ding{55} & Passively agreed and expanded; failed to probe for specifics. thinking.
\\ \cmidrule{1-4}
CollabLLM & I believe the challenge in environmental protection is reducing plastic pollution... & \ding{55} & Passively agreed and expanded; failed to probe for specifics.
\\ \cmidrule{1-4}
Sotopia-RL & You believe that environmental protection is one of the main challenges... This is a very important issue. & \ding{55} & Passively agreed and expanded; failed to probe for specifics.
\\ \cmidrule{1-4}
Qwen3-8B (base) & The challenges you mentioned are indeed very critical. Environmental protection... & \ding{55} & Passively agreed and expanded; failed to probe for specifics.
\\ \cmidrule{1-4}
IntElicit & You mentioned that environmental protection is a core challenge, but specifically which aspect? For example, technical difficulties... & \ding{51} & Prompted for specifics to refine the abstract concept.
\\ \cmidrule{1-4}
\bottomrule
\end{tabular}
\label{tab:case_perfunctory}
\end{table}

\end{document}